\documentclass[10pt,twocolumn,letterpaper]{article}

\usepackage[pagenumbers]{wacv} 

\usepackage{graphicx}
\usepackage{amsmath}
\usepackage{amssymb}
\usepackage{booktabs}
\usepackage{multirow}
\usepackage{colortbl}
\usepackage{array}
\usepackage{makecell}
\usepackage[ruled,vlined]{algorithm2e}
\usepackage{dblfloatfix}

\usepackage[pagebackref,breaklinks,colorlinks]{hyperref}

\newcommand\tab[1][1cm]{\hspace*{#1}}

\usepackage[capitalize]{cleveref}
\crefname{section}{Sec.}{Secs.}
\Crefname{section}{Section}{Sections}
\Crefname{table}{Table}{Tables}
\crefname{table}{Tab.}{Tabs.}

\def\wacvPaperID{567} 
\def\confName{WACV}
\def\confYear{2025}

\begin{document}

\title{Class-Agnostic Visio-Temporal Scene Sketch Semantic Segmentation}

\author{Aleyna Kütük \tab Tevfik Metin Sezgin\\
Department of Computer Engineering, KUIS AI Center, Koç University\\
{\tt\small \{akutuk21, mtsezgin\}@ku.edu.tr}
}
\maketitle

\begin{abstract}
   Scene sketch semantic segmentation is a crucial task for various applications including sketch-to-image retrieval and scene understanding. Existing sketch segmentation methods treat sketches as bitmap images, leading to the loss of temporal order among strokes due to the shift from vector to image format. Moreover, these methods struggle to segment objects from categories absent in the training data. In this paper, we propose a Class-Agnostic Visio-Temporal Network (CAVT) for scene sketch semantic segmentation. CAVT employs a class-agnostic object detector to detect individual objects in a scene and groups the strokes of instances through its post-processing module. This is the first approach that performs segmentation at both the instance and stroke levels within scene sketches. Furthermore, there is a lack of free-hand scene sketch datasets with both instance and stroke-level class annotations. To fill this gap, we collected the largest Free-hand Instance- and Stroke-level Scene Sketch Dataset (FrISS) that contains 1K scene sketches and covers 403 object classes with dense annotations. Extensive experiments on FrISS and other datasets demonstrate the superior performance of our method over state-of-the-art scene sketch segmentation models. The code and dataset will be made public after acceptance.
\end{abstract}

\section{Introduction}
\label{sec:intro}

\begin{figure}[t]
  \centering
   \includegraphics[width=\linewidth]{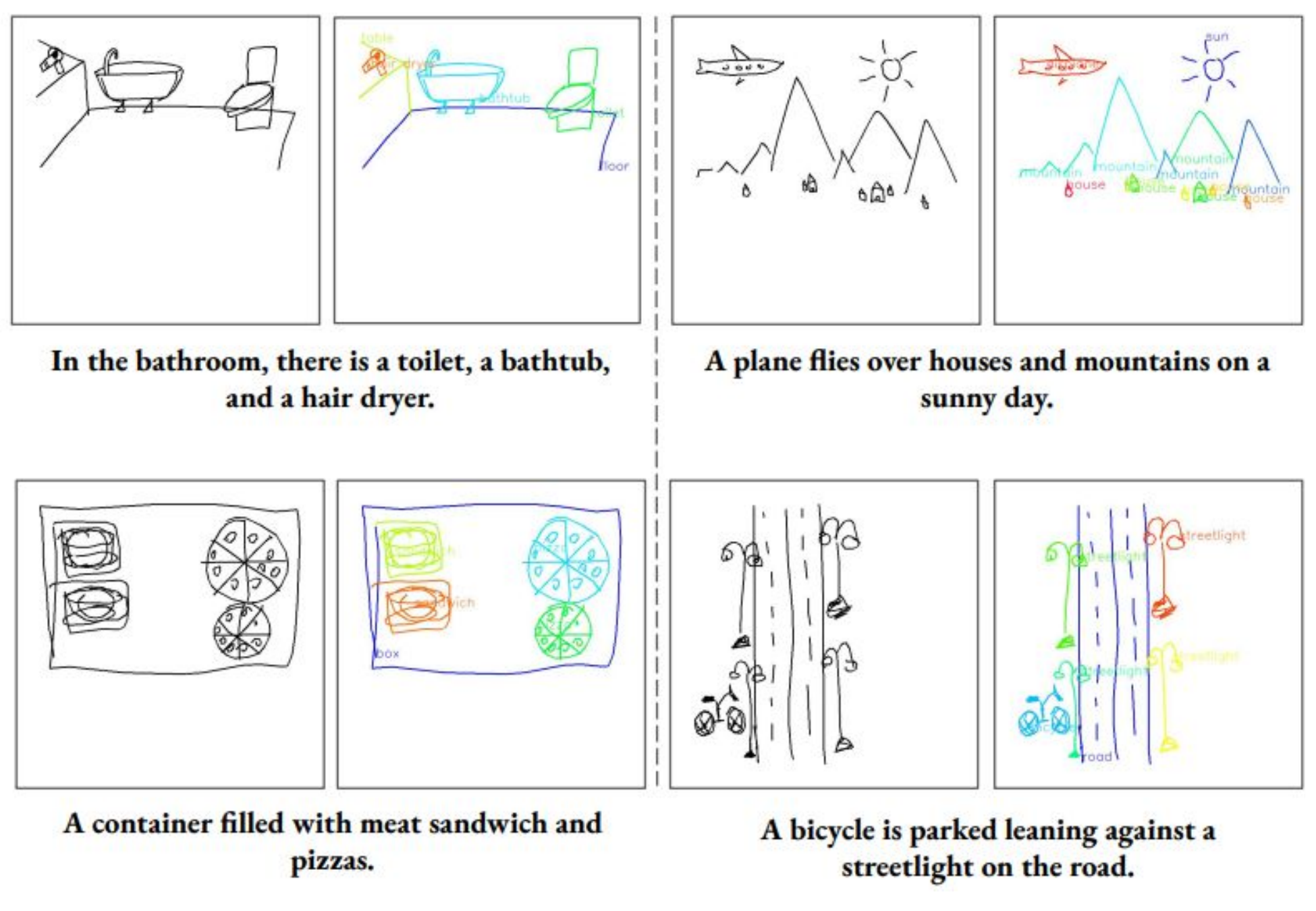}
   \caption{Sample scene sketches from FrISS dataset, each paired with corresponding textual scene descriptions. For each pair, the left image shows the black-and-white sketch, while the right image highlights the instance and stroke-level class annotations.}
  \label{friss-dataset-samples-colorful}
\end{figure}

Sketching is a rapid and widely adopted way for humans to visually express ideas. Especially with the rise of touchscreen technology, understanding hand-drawn sketches has become an essential task in the field of human-computer interaction. The field of sketch understanding includes various tasks such as sketch recognition, sketch-based image retrieval, and sketch segmentation. Sketch semantic segmentation stands out as a pivotal task, offering broad applicability in the analysis of sketches and facilitating tasks like sketch-based image retrieval. Despite the considerable attention given to semantic segmentation in natural images \cite{chen2017deeplab, badrinarayanan2017segnet, noh2015learning}, this task
remains relatively underexplored in sketches. Earlier studies on sketch segmentation have mostly concentrated on segmenting single-object sketches into semantically meaningful parts \cite{kaiyrbekov2019deep, zheng2023sketch, yang2021sketchgnn, wu2018sketchsegnet, zhu2018part, li2018fast}. On the other hand, recent attention has shifted towards scene-level sketch semantic segmentation \cite{sun2012free, zou2018sketchyscene, ge2022exploring, zhang2023stroke, bourouis2023open, yang2023scene}.

\begin{table*}
\centering
\resizebox{0.85\textwidth}{!}{
\begin{tabular}{l|cccccccc}
\hline 
\noalign{\smallskip}
Dataset & \# of Sketches & \# of Cat. & Vector & Free-hand & Scene-level & Publicly Available & Annot. Type \\
\noalign{\smallskip}
\hline
\noalign{\smallskip}
QMUL Shoe \cite{yu2016sketch} & 419 & 1 & & \checkmark & & \checkmark & C, I \\
QMUL Chair \cite{yu2016sketch} & 297 & 1 & & \checkmark & & \checkmark & C, I \\
Sketchy \cite{Sangkloy2016sketchy} & 75K & 125 & \checkmark & \checkmark & & \checkmark & C, I \\
TU-Berlin \cite{Eitz2012humans} & 20K & 250 & \checkmark & \checkmark &  & \checkmark & C, I \\
QuickDraw \cite{ha2018a} & 50M+ & 345 & \checkmark & \checkmark &  & \checkmark & C, I \\
\noalign{\smallskip}
\hline
\noalign{\smallskip}
SketchyScene \cite{zou2018sketchyscene} & 7K+ & 45 & & & \checkmark & \checkmark & C, I \\
SketchyCOCO \cite{gao2020sketchycoco} & 14K+ & 17 & & & \checkmark & \checkmark & C, I \\
SKY-Scene \cite{ge2022exploring} & 7K+ & 30 & & & \checkmark & \checkmark & C \\
TUB-Scene \cite{ge2022exploring} & 7K+ & 35 & & & \checkmark & \checkmark & C \\
CBSC \cite{zhang2018context} & 331 & 74 & \checkmark & \checkmark & \checkmark & \checkmark & C, I \\
FS-COCO \cite{chowdhury2022fs} & 10K & 92-150 & \checkmark & \checkmark & \checkmark & \checkmark & D \\
SFSD \cite{zhang2023stroke} & 12K+ & 40 & \checkmark & \checkmark & \checkmark & & C \\ 
\noalign{\smallskip}
\hline
\noalign{\smallskip}
FrISS (Ours) & 1K & 403 & \checkmark & \checkmark & \checkmark & \checkmark * & C, D, I \\
\noalign{\smallskip}
\hline
\noalign{\smallskip}
\end{tabular}}
\caption{Summary of the sketch datasets. C, I, and D denote class-level annotations, instance-level annotations, and scene sketch textual descriptions, respectively. \checkmark *: the dataset will be publicly available after acceptance.}
\label{table:dataset_comparison}
\end{table*}

Sketches are processed either as stroke sequences or bitmap images. Many methodologies treat sketches as images and address sketch segmentation similarly to image segmentation tasks \cite{sun2012free, zou2018sketchyscene, ge2022exploring, bourouis2023open, yang2023scene}. However, this direct approach often leads to the loss of temporal stroke information. As sketches consist of stroke sequences, capturing the stroke order can significantly enhance semantic segmentation performance. Moreover, current research on scene sketch segmentation mainly focuses on assigning a class to each pixel or stroke within a scene, thus segmenting scene sketches at the class level. Unfortunately, these methods cannot distinguish between individual objects that belong to the same class, such as two zebra instances in the same scene. To overcome these limitations, we introduce the Class-Agnostic Visio-Temporal Network (CAVT) that processes scene sketches and generates stroke-level groupings of instances without relying on predefined class labels. Our approach leverages visual information via an object detector and incorporates the temporal order of strokes using both a post-processing module and an RGB coloring technique.

The primary challenge for scene sketch semantic segmentation lies in the absence of large-scale scene sketch datasets. Existing scene sketch datasets are typically constructed by inserting pre-defined clip-art or free-hand single-instance sketches into the layouts of reference images \cite{zou2018sketchyscene, gao2020sketchycoco, ge2022exploring}. These datasets preserve the scene sketches in image format, limiting their utilization in stroke-based sketch methods. More recently, scene datasets have been collected by instructing participants to draw scenes based on reference natural images \cite{chowdhury2022fs, zhang2023stroke}. However, this often results in the loss of participants' natural drawing behavior, as individuals tend to replicate the object positions and postures from the reference images. 

In this work, we collected the largest Free-hand Instance- and Stroke-level Scene Sketch Dataset (FrISS), consisting of free-hand scene sketches in vector format, accompanied by textual descriptions, verbal audio recordings, and annotations at both the stroke and instance levels. To capture natural drawing behavior, participants were provided only with textual scene descriptions during the drawing process, without being shown any reference images. This approach ensures that FrISS features a diverse range of scene sketches that are not mere copies of reference images. Moreover, we avoided prolonged drawing sessions or multiple attempts, thus preventing artificially polished scene sketches. In summary, our main contributions are highlighted as follows:
\begin{enumerate}
    \item We propose CAVT, a novel scene sketch semantic segmentation pipeline, that utilizes both visual and temporal information in the scene. This is the first study on scene sketch semantic segmentation that works at both instance and stroke levels.
    \item We introduce FrISS, a densely annotated dataset that includes 1K free-hand scene sketches covering 403 object categories. FrISS can promote future stroke-based scene-level studies.
     \item We conduct extensive experiments on FrISS and other free-hand scene sketch datasets and show that our approach achieves state-of-the-art performance.
\end{enumerate}

\section{Related Work}
\label{sec:relatedwork}

\subsection{Sketch Semantic Segmentation}

Existing works on sketch semantic segmentation mostly focus on single-object sketch datasets and divide an object into its semantically valid parts \cite{kaiyrbekov2019deep, zheng2023sketch, yang2021sketchgnn, wu2018sketchsegnet, zhu2018part, li2018fast}. On the other hand, scene-level sketch semantic segmentation aims to distinguish individual object instances within the scene. Regarding the processing of sketches, these studies can be divided into two main groups: image-based and sequence-based. Image-based methods typically treat sketches as raster images and output pixel-level segmentation predictions; whereas sequence-based methods utilize stroke-level information and assign semantic labels to each stroke in a sketch. Even if the majority of studies on single-object sketch semantic segmentation lie in the sequence-based methods \cite{kaiyrbekov2019deep, zheng2023sketch, yang2021sketchgnn, wu2018sketchsegnet}, there are not many studies conducted on stroke-level scene sketch semantic segmentation. This is mostly due to the lack of large-scale scene sketch datasets with stroke-level class annotations.

\begin{figure*}[ht!]
  \centering
  \includegraphics[width=0.75\textwidth]{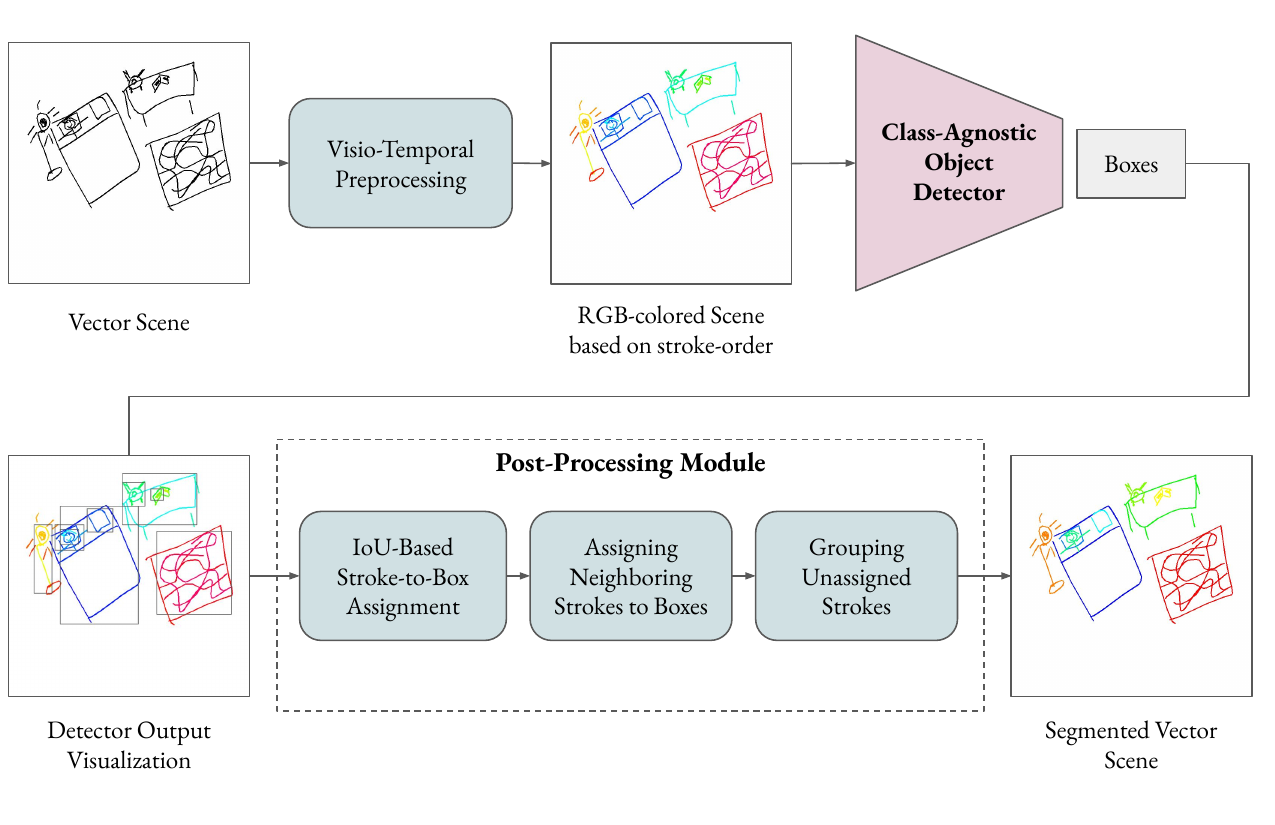}
  \caption{The overall pipeline of CAVT}
  \label{fig:cavt_pipeline}
\end{figure*}

Prior works on scene sketch semantic segmentation treat the task as a semantic image segmentation problem, disregarding the stroke order \cite{zou2018sketchyscene, ge2022exploring, bourouis2023open, yang2023scene}. SketchyScene \cite{zou2018sketchyscene} is the pioneering study that assigns object categories at the pixel level. Ge \etal~\cite{ge2022exploring} proposed a deep-shallow feature fusion network based on DeepLab-v2 \cite{chen2017deeplab}, examining the influence of local details on scene sketch segmentation. Bourouis \etal~\cite{bourouis2023open} introduced the first language-supervised scene sketch segmentation method by utilizing sketch captions. In contrast, Zhang \etal~\cite{zhang2023stroke} developed an RNN-GCN-based architecture, marking the first stroke-level approach to scene sketch semantic segmentation. Their study is the most relevant to ours since they also utilize visual, sequential, and spatial information on stroke sequences. However, their code is not publicly available for comparison.

Scene sketch segmentation works mostly focus on assigning each pixel or stroke to a specific class in a given scene. Therefore, different objects belonging to the same category cannot be distinguished at the instance level. In contrast, we propose a novel class-agnostic scene segmentation pipeline that can differentiate object instances in a given scene, regardless of their classes.

\subsection{Sketch Datasets}

Sketch datasets can be categorized into two primary types: single-object and scene sketch datasets. Single-object sketch datasets feature one object instance per sketch, while scene sketch datasets encompass drawings with multiple objects. Table \ref{table:dataset_comparison} provides a summary of the sketch datasets and our proposed scene sketch dataset, FrISS.

QMUL Shoe \cite{yu2016sketch}, QMUL Chair \cite{yu2016sketch}, and Sketchy \cite{Sangkloy2016sketchy} are multi-modal single-object sketch datasets that contain corresponding natural images paired with each sketch. TU Berlin \cite{Eitz2012humans} is the first large-scale free-hand single-object sketch dataset, that is collected via crowdsourcing. QuickDraw \cite{ha2018a} is the largest free-hand single sketch dataset, and it is gathered through an online game.

A growing number of large-scale scene-level sketch datasets have been proposed due to the importance of higher-level sketch understanding. SketchyScene \cite{zou2018sketchyscene} pioneered this field, assembling clip art-like single-object sketches onto reference images as layout templates. SketchyCOCO \cite{gao2020sketchycoco} is another synthetically generated scene sketch dataset that integrates free-hand single-object datasets into the corresponding mask area of COCO-Stuff \cite{caesar2018coco} real images. Ge \etal~\cite{ge2022exploring} introduced two more semi-synthetic scene datasets, called as SKY-Scene and TUB-Scene. Although synthetic scene sketch data generation offers a quick solution to the scarcity of large-scale scene datasets, it lacks the authenticity of human drawing behavior. Moreover, none of these synthetic datasets are available in vector storage formats, rendering them unsuitable for our stroke-based approach. FS-COCO \cite{chowdhury2022fs} stands out as the first free-hand scene sketch dataset collected in vector format, accompanied by scene captions. However, it lacks stroke- or object-level annotations, hindering semantic segmentation experiments. SFSD \cite{zhang2023stroke} is another free-hand scene sketch dataset, offering both vector storage format and stroke-level class annotations, but it is not publicly available. Lastly, CBSC \cite{zhang2018context} emerges as the sole publicly accessible free-hand scene sketch dataset with instance-level class annotation in the vector storage format. Thus, we leverage CBSC to test our network. To address the lack of free-hand scene sketch datasets, we introduce FrISS, which contains free-hand scene sketches annotated at both instance and stroke levels.

\section{Methodology}
\label{sec:methodology}

In this section, the architecture of CAVT and the generation process of its training dataset are explained. As seen in Figure \ref{fig:cavt_pipeline}, CAVT consists of two sub-modules: (i) the Class-Agnostic Visio-Temporal object detector and (ii) the Post-Processing module. First, each scene sketch is pre-processed using an RGB coloring technique to preserve the temporal stroke order. These color-coded sketches are then passed through the Class-Agnostic Object Detector to generate prediction boxes. Subsequently, the Post-Processing module refines the detector's outputs using a set of rules for stroke-level instance grouping by leveraging temporal stroke order and spatial features. Finally, CAVT produces stroke groups belonging to object instances in the scene.

\subsection{Class-Agnostic Visio-Temporal Detector}
\label{sec:cavt-detector}

To proceed with an appropriate object detector, we investigated the cross-domain object detection studies \cite{deng2021unbiased, jiang2021decoupled, topal2023domainadaptive} in the literature. DASS-Detector \cite{topal2023domainadaptive} leverages YOLOX \cite{ge2021yolox} and stands out for its high performance within its domain. Inspired by their work, we also utilize  YOLOX in our study. Fully-supervised detectors are typically trained to recognize specific predefined classes, restricting their ability to detect objects beyond these predetermined categories. To address this constraint, YOLOX is trained in a class-agnostic manner, in which the detector solely predicts potential object areas without the need for classification. We conduct an ablation study to evaluate the impact of our approach and discuss it in Sec. \ref{sec:ablation}. Our trained detector offers predictions concerning potential object regions within sketch scenes. These predictions solely approximate object-bounding boxes on the coordinate plane. Therefore, we introduce a post-processing module designed to group object strokes by leveraging the bounding box predictions.

\subsection{Post-Processing Module}
\label{sec:postprocessing}

This module performs stroke-level segmentation for individual sketches by utilizing the output from the object detector. The full algorithm for the post-processing module is provided in Algorithm \ref{alg:postprocessing_algorithm} in the Supplementary Material. The steps involved in this module are as follows:

\begin{enumerate}
    \item The predicted bounding boxes are sorted in ascending order based on their area, from smallest to largest.
    \item \textit{IoU-Based Stroke-to-Box Assignment:} Starting with the smallest bounding box, the stroke sequence with the highest Intersection over Union (IoU) compared to the selected box is identified. If the IoU value surpasses a threshold called \textit{IoU\_threshold}, the corresponding stroke set is assigned to that bounding box.
    \item \textit{Assigning Neighboring Strokes to Boxes:} The unassigned strokes are then evaluated based on their overlap ratio. For each of the remaining longest stroke sequences, if the overlap ratio between the sequence and the nearest bounding box exceeds a threshold called \textit{OR\_threshold}, the stroke set is assigned to that box. The overlap ratio is calculated by dividing the area of intersection between the bounding box and the stroke set by the total area of the stroke set.
    \item \textit{Grouping Unassigned Strokes:} Strokes that remain unassigned to any bounding box after these steps are considered separate objects, and their coordinates are added to the list of predicted boxes.
    \item The coordinates of each bounding box are updated based on the latest stroke assignments. Each box's dimensions are adjusted to become the smallest bounding box enclosing its assigned stroke set.
    \item These steps are repeated until no further changes occur in stroke groupings, ensuring that each stroke is assigned to a corresponding object bounding box.
\end{enumerate}

Both the \textit{IoU\_threshold} and \textit{OR\_threshold} are determined using a grid-search algorithm (see in Supplementary Material Sec. \ref{sec:details_postprocessing}). The object detector produces bounding boxes without class predictions, so strokes are grouped without class information. This enables the utilization of an external sketch object classifier, offering several advantages: (1) Both stroke- and image-based single sketch classifiers can be employed, each capable of identifying broader or narrower object categories, or sketches with varying complexities; (2) Inference time and required memory can be adjusted based on the chosen classifiers.

\subsection{Synthetic Dataset Preparation for Training}
\label{sec:training-dataset-preparation}

Object detection models are widely used in the literature \cite{zhang2022dino, li2023yolov6, xu2021end}. However, their direct application to the sketch domain faces challenges due to the domain shift from real-life images to scene sketches. Achieving fully supervised detector training on sketches necessitates a large-scale instance-level scene sketch dataset. Furthermore, the training dataset should maintain strokes in vector storage format to utilize temporal cues effectively. Unfortunately, none of the existing large-scale datasets offer both instance-level annotation and vector storage format \cite{zou2018sketchyscene, gao2020sketchycoco, ge2022exploring, chowdhury2022fs}.

\begin{table*}
\centering
\resizebox{0.9\textwidth}{!}{\begin{tabular}{l|c|ccccc|ccccc|cccc}
\noalign{\smallskip}
\hline
\noalign{\smallskip}
\multirow{2}{*}{Dataset} & \multirow{2}{*}{Cats.} & & \multicolumn{3}{c}{Category per sketch} & & & \multicolumn{3}{c}{Objects per sketch} & & & \multicolumn{3}{c}{Strokes per sketch} \\ \cline{4-6} \cline{9-11} \cline{14-16} & & & Max & Min & Mean & & & Max & Min & Mean & & & Max & Min & Mean \\
\noalign{\smallskip}
\hline
\noalign{\smallskip}
SketchyScene \cite{zou2018sketchyscene} & 45 & & 19 & 13 & 6.88 & & & 94 & 3 & 16.71 & & & - & - & - \\
SketchyCOCO \cite{gao2020sketchycoco} & 17 & & 6 & 1 & 2.33 & & & 35 & 2 & 10.93 & & & - & - & - \\
CBSC \cite{zhang2018context} & 74 & & 10 & 3 & 4.23 & & & 16 & 3 & 4.72 & & & 185 & 6 & 33.14 \\
FS-COCO \cite{chowdhury2022fs} & 92 / 150 & & 5 / 25 & 1 / 1 & 1.37 / 7.17 & & & - & - & - & & & 561 & 5 & 75.86 \\
SFSD \cite{zhang2023stroke} & 40 & & 11 & 1 & 4.46 & & & 43 & 2 & 7.76 & & & 699 & 9 & 146.64 \\
\noalign{\smallskip}
\hline
\noalign{\smallskip}
FrISS (Ours) & 403 & & 10 & 1 & 4.33 & & & 30 & 1 & 6.04 & & & 186 & 4 & 35.81 \\
\noalign{\smallskip}
\hline
\noalign{\smallskip}
\end{tabular}}
\caption{Comparison and statistics of scene sketch datasets}
\label{table:datasets_statistics}
\end{table*}

To train an object detector for the sketch domain, we created a large-scale, synthetically generated scene sketch dataset. To ensure our object detector's robustness across various categories and drawing styles, we utilized QuickDraw \cite{ha2018a}, which offers a wide range of categories and diverse sketch styles. Each scene is composed of a minimum of 2 and a maximum of 8 randomly chosen objects from a pool of 345 categories, with 70K drawing instances per category. Objects are randomly scaled to have a large side length ranging from 50 to 700 pixels and positioned randomly within the scene. To prevent extreme overlapping between objects, we ensure that the intersection-over-union (IOU) value between them remains below 0.35. Scenes are created in two potential sizes: 720x1280 or 1280x720 pixels. To capture the temporal order, each stroke is assigned a color from a spectrum spanning blue to red based on its order (see Supplementary Material Sec. \ref{sec:details_rgbcoloring}). In total, we generated 11.5K synthetic drawing scenes under these settings, allocating 10K for training and 1.5K for validation.

\section{The FrISS Dataset}
\label{sec:dataset}

We propose the largest Free-hand Instance- and Stroke-level Scene sketch dataset (FrISS) that includes scene sketches in vector format, stroke-level class and instance annotations, sketch-text pairs, and verbal audio clips paired with each scene. The data construction process involves two primary stages: (i) sketch collection and (ii) sketch annotation. This section elaborates on these stages and provides statistics and analysis on the FrISS dataset.

\begin{figure}[th!]
  \centering
   \includegraphics[width=\linewidth]{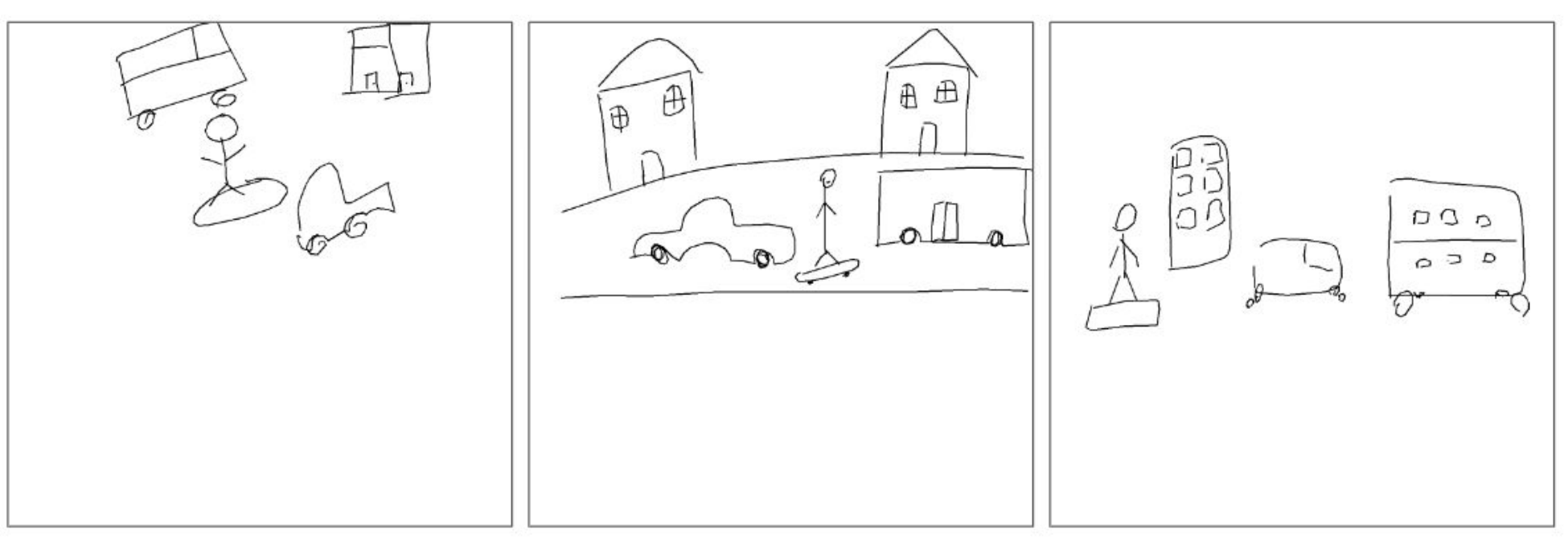}
  \caption{Sample scenes taken from FrISS that are drawn by three individuals by referring to the same textual scene description}
  \label{same-description-samples}
\end{figure}

\subsection{Sketch Collection}
\label{sec:sketch_collection}

We developed a web application to collect scene sketches, following similar data collection methods as in previous studies \cite{chowdhury2022fs, zhang2023stroke}. Visuals of the web application are provided in Supplementary Material Sec. \ref{sec:webui_friss}. We recruited 100 volunteer participants with varying levels of drawing skills, each tasked with creating 10 distinct scene sketches based on textual scene descriptions. The textual scene descriptions provided during the drawing phase are either sourced from captions within the MS COCO dataset \cite{lin2014microsoft} or constructed by us. Details on the generation of scene descriptions, along with examples, are provided in Supplementary Material Sec. \ref{sec:descriptions_friss}. To avoid influencing participants with predefined layouts or poses, no visual references were provided. As shown in Figure \ref{same-description-samples}, the arrangement and diversity of objects in the scenes varied significantly when participants sketched scenes without visual guidance.

Each participant was given 1.5 minutes to complete each scene. The time limit was determined through pilot studies with a group of volunteers. These studies showed that a shorter time often led to incomplete drawings, while a longer time resulted in excessively detailed sketches. Participants were allowed to redraw objects within the time limit but were not permitted to restart the scene with extra time. Allowing multiple attempts could lead to unrealistically polished sketches. Additionally, participants were asked to verbally describe their scenes as they drew. To ensure comfort and clarity, they were encouraged to speak in their native language. The verbal explanations were recorded during the drawing process, enabling FrISS to support research on tasks such as speech-based sketch studies.

\subsection{Sketch Annotation}

In the second phase of data collection, participants were presented with scenes they had previously drawn. They annotated each stroke with both instance and category information. Figure \ref{friss-dataset-samples-colorful} shows sample sketch-text pairs from the FrISS dataset and their colored annotations. Different colors are used to visualize instance-level annotations of the objects from the same category (e.g., pizzas, mountains).

To avoid interrupting the natural drawing process, we collected sketch annotations separately from the drawing phase. This phase was conducted under our supervision to ensure accurate annotations. Each stroke in a scene was assigned to its corresponding object category, with incomplete or ambiguous strokes labeled as \textit{'incomplete'} and subsequently excluded from the scene. Additionally, we manually reviewed the annotations for accuracy and assessed the quality of the scenes. Any mislabeled object strokes were either corrected or eliminated from the dataset.

\subsection{Statistics and Analysis}
\label{sec:datasetstatistics}

Table \ref{table:datasets_statistics} provides a statistical comparison of various scene sketch datasets, focusing on category, object, and stroke counts per sketch. Our dataset covers a wider range of object categories compared to previous scene datasets. Additionally, each scene sketch was collected within a 1.5-minute timeframe, resulting in simpler sketches resembling participants' daily drawings. Other free-hand scene sketch datasets \cite{chowdhury2022fs, zhang2023stroke} allow more time for drawings and multiple drawing attempts, which results in extremely detailed scene sketches. On the other hand, our scene sketches contain an average of approximately 36 strokes per scene, significantly fewer than other datasets in terms of complexity. Refer to Figure \ref{friss-dataset-samples-colorful} for sample scenes from our FrISS dataset. Thus, FrISS stands out by including both instance- and stroke-level class annotations. Additional scene samples and comparisons are available in Supplementary Material Sec. \ref{sec:additional_details_friss}.

\section{Experiments}
\label{sec:experiments}

\subsection{Datasets}

We utilize temporal stroke information in our pipeline, thus it limits the range of applicable datasets for evaluation. Therefore, we assessed our approach using only the test partitions of FrISS and CBSC \cite{zhang2018context}. FrISS comprises 1K free-hand scene sketches spanning 403 object categories, with 236 categories overlapping with the QuickDraw classes \cite{ha2018a}. We reserved 500 scene sketches for testing, while the remaining sketches were divided into validation (145 sketches) and training (355 sketches) sets. CBSC dataset consists of 222 free-hand scene sketches in its test partition, covering 74 object categories and these categories fully align with QuickDraw, except for the \textit{'person'} class. However, the visual characteristics of the \textit{'yoga'} class of the QuickDraw closely resemble those of the \textit{'person'} class in other scene sketch datasets. Therefore, we map the \textit{'person'} class to \textit{'yoga'} class during the evaluation. 

\subsection{Sketch Classification}
\label{sec:post_classification}

As discussed in Sec. \ref{sec:methodology}, CAVT generates segmented stroke groups without any category assignments. Thus, we utilized one stroke-based and one image-based sketch classifier. First, we investigated the performances of state-of-the-art stroke-based sketch classifiers \cite{xu2021multigraph, ha2017neural, ribeiro2020sketchformer}. Since Sketchformer \cite{ribeiro2020sketchformer} achieves superior performance, it was selected as the external classifier for categorizing sketches segmented by CAVT. Secondly, we trained various CNN-based classifiers using the training sets of QuickDraw and FrISS. Among these, Inception-V3 \cite{szegedy2016rethinking}
outperforms others. Hence, we further utilize our trained Inception-V3 as a second external classifier. In the following sections, we call the end-to-end CAVT + Sketchformer pipeline as \textit{CAVT-S}, and CAVT + pre-trained-Inception-V3 pipeline as \textit{CAVT-I}. A detailed analysis of classifiers can be found in Supplementary Material Sec. \ref{sec:external_classifiers}. 

\subsection{Evaluation Metrics}

Earlier works utilize metrics that are commonly used to evaluate image segmentation models. Hence, we follow the standard four metrics that are used in our competitor models  \cite{bourouis2023open, ge2022exploring, zou2018sketchyscene} for fair comparison. These metrics are listed as follows: Overall Pixel Accuracy (OVAcc), Mean Pixel Accuracy (MeanAcc), Mean Intersection over Union (MIoU), and Frequency Weighted Intersection over Union (FWIoU). Still, there is no available metric specifically designed for stroke-level scene sketch semantic segmentation. Thus, we propose two additional metrics for stroke-level evaluation:

\begin{itemize}
    \item \textbf{All or Nothing (AoN):} evaluates the ratio of correctly predicted stroke groups. If a single stroke of an object is mislabeled, then the result becomes incorrect.
    \item \textbf{Stroke-level Intersection over Union (S-IoU):} calculates the largest overlap ratio of the actual and the predicted stroke groups, and averages the overlap ratio for all ground truth stroke groups.
\end{itemize}

Our competitor models perform class-level segmentation and require bitmap images as input. Therefore, we could not compare our results with earlier works on these metrics.

\begin{table*}
\centering
\def\arraystretch{0.9}
\begin{tabular}{lc|cccccccccc}
\noalign{\smallskip}
\hline
\noalign{\smallskip}
\multirow{2}{*}{Model} & & & \multicolumn{4}{c}{CBSC-SKY} & & \multicolumn{4}{c}{CBSC-SS} \\ \cline{4-7} \cline{9-12} \noalign{\smallskip} & & & OVAcc & MeanAcc & MIoU & FWIoU & & OVAcc & MeanAcc & MIoU & FWIoU \\ 
\noalign{\smallskip}
\hline
\noalign{\smallskip}
LDP & & & 54.56 & 52.82 & 33.47 & 37.96 & & 47.85 & 36.17 & 23.81 & 32.93 \\
CAVT-S & & & 70.24 & 73.89 & 51.21 & 59.22 & & 71.25 & 73.29 & 51.92 & 60.30 \\ 
CAVT-I & & & \textbf{73.76} & \textbf{74.08} & \textbf{53.38} & \textbf{61.89} & & \textbf{73.13} & \textbf{75.26} & \textbf{52.45} & \textbf{60.56} \\ 
\noalign{\smallskip}
\hline
\noalign{\smallskip}
\noalign{\smallskip}
\hline
\noalign{\smallskip}
\multirow{2}{*}{Model} & & & \multicolumn{4}{c}{FrISS-SKY} & & \multicolumn{4}{c}{FrISS-SS} \\ \cline{4-7} \cline{9-12} \noalign{\smallskip} & & & OVAcc & MeanAcc & MIoU & FWIoU & & OVAcc & MeanAcc & MIoU & FWIoU \\ 
\noalign{\smallskip}
\hline
\noalign{\smallskip}
LDP & & & 44.33 & 27.24 & 14.91 & 31.89 & & 41.17 & 29.97 & 15.09 & 27.82 \\
CAVT-S  & & & 65.39 & \textbf{62.33} & \textbf{34.88} & \textbf{54.86} & & 60.02 & \textbf{60.11} & \textbf{33.09} & 48.11 \\ 
CAVT-I  & & & \textbf{66.56} & 62.08 & 34.18 & 54.40 & & \textbf{61.54} & 55.07 & 31.83 & \textbf{48.19} \\ 
\noalign{\smallskip}
\hline
\noalign{\smallskip}
\end{tabular}
\caption{Comparison of CAVT against LDP \cite{ge2022exploring} on the CBSC-SS CBSC-SKY, FrISS-SS, and FrISS-SKY datasets.}
\label{table:cavt_results_ldp}
\end{table*}
\begin{table*}
\centering
\def\arraystretch{0.9}
\resizebox{\textwidth}{!}{%
\begin{tabular}{l|ccccccccccccccc}
\noalign{\smallskip}
\hline
\noalign{\smallskip}
\multirow{2}{*}{Model} & & \multicolumn{4}{c}{CBSC} & & \multicolumn{4}{c}{FrISS-QD} & & \multicolumn{4}{c}{FrISS} \\ \cline{3-6} \cline{8-11} \cline{13-16}\noalign{\smallskip} & & OVAcc & MeanAcc & MIoU & FWIoU & & OVAcc & MeanAcc & MIoU & FWIoU & & OVAcc & MeanAcc & MIoU & FWIoU\\
\noalign{\smallskip} 
\hline
\noalign{\smallskip}
OV & & 62.64 & 62.94 & 45.15 & 49.34 & & 64.66 & 54.67 & 38.14 & 50.68 & & 41.13 & 41.84 & 25.41 & 29.92 \\
CAVT-S* & & 81.21 & 81.87 & 68.71 & 70.13 & & 80.90 & \textbf{76.99} & 64.95 & 69.53 & & - & - & - & -  \\
CAVT-I* & & \textbf{83.52} & \textbf{82.36} & \textbf{71.97} & \textbf{73.14} & & \textbf{81.89} & 75.50 & \textbf{65.81} & \textbf{70.97} & & \textbf{72.71} & \textbf{46.46} & \textbf{37.17} & \textbf{58.05} \\
\noalign{\smallskip}
\hline
\noalign{\smallskip}
\end{tabular}
}
\caption{Comparison of CAVT with Open Vocabulary (OV) \cite{bourouis2023open} on the CBSC \cite{zhang2018context}, FrISS-QD, and FrISS datasets.}
\label{table:cavt_results_ov}
\end{table*}

\subsection{Implementation Details}
\label{sec:implementationdetails}

The sole trainable component of our network is the Class-Agnostic Visio-Temporal Object Detector, built upon the YOLOX framework \cite{ge2021yolox}. During model training, we employed the MMDetection library, training YOLOX with default configurations while modifying only the total number of categories to 1. Our training process utilizes a single Tesla T4 GPU with a batch size of 16, spanning 600 epochs. We compared our results with the Local Detail Perception (LDP) \cite{ge2022exploring} and the Open Vocabulary (OV) \cite{bourouis2023open}. However, when comparing CAVT with these models, several adjustments to the datasets and our evaluation process are necessary:

\begin{itemize}
    \item LDP is trained on categories from SKY-Scene \cite{ge2022exploring} and SketchyScene \cite{zou2018sketchyscene}. Additionally, we use Sketchformer \cite{ribeiro2020sketchformer} as our sketch classifier, which only supports the 345 categories from QuickDraw \cite{ha2018a}. To ensure a fair comparison, we created five distinct sub-datasets: \textit{FrISS-SKY} and \textit{CBSC-SKY} include objects from the common classes shared between QuickDraw, SKY-Scene, and FrISS/CBSC; \textit{FrISS-SS} and \textit{CBSC-SS} feature objects from the common categories of QuickDraw, SketchyScene, and FrISS/CBSC; \textit{FrISS-QD} comprises objects from the common classes of FrISS and QuickDraw.

    \item The OV model operates without relying on pixel or stroke-level annotations, instead, it uses sketch-caption pairs. During inference, captions are generated by concatenating ground truth object categories, and OV predicts the correct class label from the given set of object classes. To ensure a fair comparison with OV, we developed alternative versions of our pipelines (CAVT-S* and CAVT-I*) that restrict the possible object classes to those present in the ground truth scene.
    
\end{itemize}

\begin{figure*}[th!]
  \centering
   \includegraphics[width=0.9\linewidth]{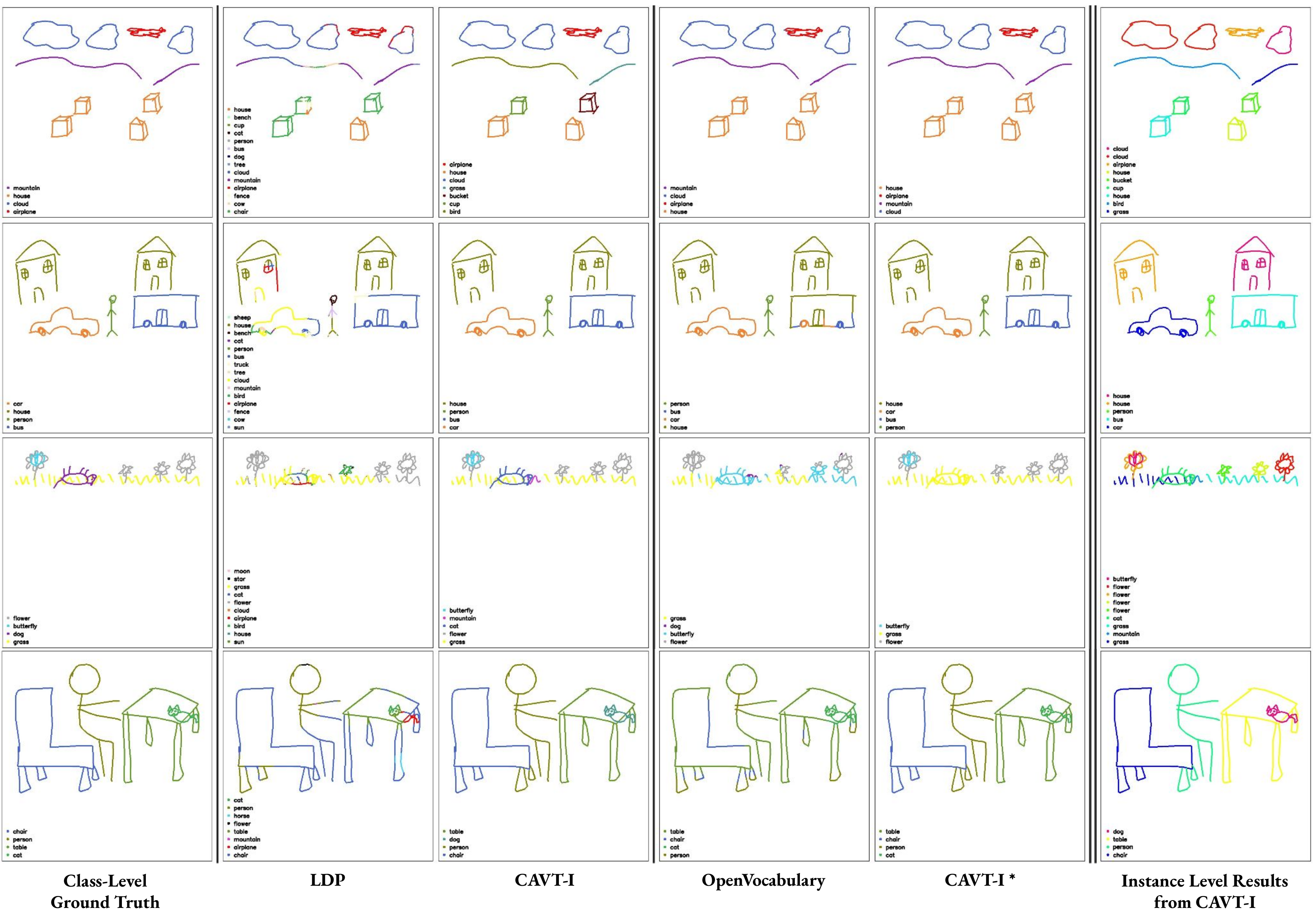}
   \caption{Visual comparison of our method with LDP \cite{ge2022exploring} and OV \cite{bourouis2023open} models that are evaluated on the FrISS-SS dataset.}
  \label{visual-results}
\end{figure*}

\begin{table*}
\centering
\resizebox{\textwidth}{!}{%
\begin{tabular}{ccc|cccccccccccccccc}
\noalign{\smallskip}
\hline
\noalign{\smallskip}
\multicolumn{3}{c}{Components:} & \multicolumn{6}{|c}{CBSC} & & \multicolumn{6}{c}{FrISS} & & \multicolumn{2}{c}{FrISS$^{sub}$} \\ \noalign{\smallskip}
\cline{4-9} \cline{11-16} \cline{18-19} \noalign{\smallskip} 
T & CA & PP & AoN & S-IoU & OVAcc & MAcc & MIoU & FWIoU & & AoN & S-IoU & OVAcc & MAcc & MIoU & FWIoU & & AoN & S-IoU \\
\noalign{\smallskip}
\hline
\noalign{\smallskip}
 & & & 
 39.80 & 68.76 & 39.54 & 38.21 & 23.74 & 28.53 & & 
 28.30 & 57.49 & 25.08 & 14.34 &  6.17 & 17.32 & & 
 23.85 & 58.23 \\
 
 & & \checkmark & 
 48.95 & 73.57 & 48.83 & 46.68 & 30.70 & 37.05 & & 
 33.40 & 60.07 & 29.77 & 18.21 &  8.83 & 20.74 & & 
 26.11 & 58.37 \\
 
 & \checkmark & \checkmark & 
 58.64 & 81.09 & 52.00 & 51.67 & 32.57 & 39.55 & & 
 47.09 & \textbf{72.89} & 32.89 & \textbf{22.33} & 9.98 & 23.17 & & 
 39.98 & \textbf{71.67} \\

\checkmark & \checkmark & \checkmark & 
\textbf{68.68} & \textbf{84.77} & \textbf{60.09} & \textbf{57.50} & \textbf{38.81} & \textbf{47.96} & & 
\textbf{51.57} & 72.77 & \textbf{36.55} & 22.04 & \textbf{10.20} & \textbf{26.12} & & 
\textbf{41.62} & 71.34 \\
\noalign{\smallskip}
\hline
\noalign{\smallskip}
\end{tabular}
}
\caption{Ablation study for the impact of including or excluding three components: temporal stroke order (T), class-agnostic training (CA), and post-processing steps (PP). FrISS$^{sub}$: calculates metrics for a subset of categories in FrISS that are not part of QuickDraw \cite{ha2018a}. The metrics \textit{OVAcc}, \textit{MeanAcc}, \textit{MIoU}, and \textit{FWIoU} are evaluated using CAVT-I, since it also supports the complete class set of FrISS.}
\label{table:ablation_cavt_components}
\end{table*}

\subsection{Comparison Against State-of-the-art (SOTA)}
\label{sec:comparisons}

The comparison results of our model with prior works on the different subsets of CBSC and FrISS datasets are shown in Tables \ref{table:cavt_results_ldp} and  \ref{table:cavt_results_ov}. Across all datasets and metric variations, under identical conditions, the gap between LDP and CAVT-S or CAVT-I is consistently between 15\% - 39\%, but it narrows to 6\% - 31\% with OV. Still, our pipeline outperforms previous SOTA by a significant margin.

Figure \ref{visual-results}  shows the qualitative comparison between our method, LDP, and OV models. Our pipeline leverages stroke information and does not assign different class labels to any point in a single stroke. This allows us to generate more coherent segmentation outputs. Moreover, we share our instance-level visual results in the rightmost column of the figure. Different from the SOTA models, we can segment different instances from the same category (2nd and 3rd rows). We provide additional visual comparisons in Supplementary Material.

\subsection{Ablation Study}
\label{sec:ablation}

In this experiment, we examine the individual effects of each component of CAVT. The key components include the use of temporal stroke order, class-agnostic training, and the post-processing module. To evaluate the impact of the post-processing module, we implement a simple stroke grouping technique as a baseline for comparison. In this method, each stroke is assigned to the nearest predicted bounding box, and the strokes assigned to the same box are grouped as a single object. 

Table \ref{table:ablation_cavt_components} illustrates the impact of each component on segmentation performance, with each one contributing a notable improvement. While the most significant component in CBSC is PP with a 7.48\% average performance increase, CA has the least effect with a 4.96\% increase. On the other hand, CA has the most effect on FrISS with an average of 6.22\% performance enhancement, while T provides the least increase with 1.82\% on average.

As detailed in Section \ref{sec:cavt-detector}, the object detector is trained using a synthetic dataset derived from QuickDraw classes \cite{ha2018a}. We excluded objects belonging to QuickDraw categories from the FrISS dataset and denoted as FrISS$^{sub}$. Later, we calculated AoN and S-IoU on this subset to evaluate the generalizability of CAVT to instances from unseen classes. Although the AoN score drops by approximately 10\%, the decrease in S-IoU remains only around 1.5\%. This indicates that CAVT can still generalize to sketch objects from unseen classes with minimal performance loss.

\section{Conclusion}
\label{sec:conclusion}
In this work, we proposed a novel pipeline for the scene sketch semantic segmentation task that identifies individual object instances at both stroke- and instance levels. We utilized both temporal information and the visual appearance of the sketches within a scene. Our approach allows us to assign a class label to each object instance without being constrained by a predefined category list. Furthermore, we introduced the FrISS dataset, comprising instance and stroke-level class annotations, sketch-text pairs, and verbal audio clips paired with each scene. We hope that FrISS facilitates a wide range of studies, including stroke-level scene sketch segmentation, speech-based sketch applications, and cross-modal research utilizing sketch-text pairs. Benefitting from FrISS, we conducted extensive experiments to show that our novel approach outperforms the state-of-the-art methods, yielding more coherent visual results in scene sketch semantic segmentation.

\section{Acknowledgement}

This study is written as a part of a Research Project supported by grants from the Scientific and Technological Research Council of Turkey (TUBITAK), Turkey (Project No. 120E489). We are also thankful for the support of the council and KUIS AI Center.

{\small
\bibliographystyle{ieee_fullname}
\bibliography{main}
\pagebreak
}

\setcounter{equation}{0}
\setcounter{figure}{0}
\setcounter{table}{0}
\setcounter{section}{0}
\makeatletter
\renewcommand{\theequation}{S\arabic{equation}}
\renewcommand{\thefigure}{S\arabic{figure}}
\renewcommand{\thesection}{S\arabic{section}}

\begin{center}
    {\Large\bfseries Class-Agnostic Visio-Temporal Scene Sketch Semantic Segmentation \\ - Supplementary Material -}
\end{center}

The Supplementary Material is organized as follows. The details regarding the Post-Processing Module of CAVT are provided in Section \ref{sec:details_postprocessing}. RGB Coloring Technique is detailed in Section \ref{sec:details_rgbcoloring}. Additional analysis on external classifiers is provided in Section \ref{sec:external_classifiers}. Additional visual results are shared for scene sketch segmentation in Section \ref{sec:additional_segmentation_results}. Lastly, additional analysis and discussions regarding to FrISS dataset and UI of data collection web application are shared in Section \ref{sec:additional_details_friss}.

\section{Details on Post-Processing Module}
\label{sec:details_postprocessing}

\subsection{Hyperparameter Optimization}

\begin{algorithm}[ht!]
\caption{Post-Processing Module}
\label{alg:postprocessing_algorithm}
\KwIn{boxes, IoU\_threshold, OR\_threshold}
\KwOut{segmented stroke groups}

\While{there is alternation in stroke grouping}{
Mark all strokes as unassigned. \\
Sort the boxes by area in ascending order.

\For{each box $b_i$ in boxes}{
    Find the longest stroke sequence $S$ that has the highest IoU with the box $b_i$. \\
    \If{the overlap ratio between $S$ and $b_i$ is more than IoU\_threshold}{
        Assign stroke sequence $S$ to bounding box $b_i$.
    }
}
\For{each unassigned longest stroke sequence $S_u$}{
    Find the nearest bounding box $b_i$. \\
    \If{the overlap ratio between $S_u$ and $b_i$ is more than OR\_threshold}{
        Assign strokes in $S_u$ to bounding box $b_i$.
    }
}
\For{each longest stroke sequence $S_u$ that are unassigned}{
   Find the boundaries $S_u$: $x\_min$, $y\_min$, $x\_max$, $y\_max$. \\
   Define a new box $b_{new}$ from values $x\_min$, $y\_min$, $x\_max$, $y\_max$. \\
   Append $b_{new}$ to the boxes. \\
   Assign each stroke in $S_u$ to $b_{new}$.
}
\For{each box $b_i$ in boxes}{
   Update the coordinates of each $b_i$ according to the most recent assignment of strokes.
}}
\end{algorithm}

The complete algorithm for the post-processing module is outlined in Algorithm \ref{alg:postprocessing_algorithm}. Furthermore, we provide details of our grid-search approach used to determine the optimal hyperparameter combination for the post-processing module. We evaluated the AoN and S-IoU scores on the validation sets of both CBSC and FrISS and selected the top-performing parameter combination based on the average of all scores. Table \ref{table:ablation_cavt_postprocessing} presents the results for the top-performing parameter combination. The parameters in the ablation study are explained as follows:

\begin{table*}
\centering
\resizebox{\textwidth}{!}{\begin{tabular}{c|cccc|ccccccc|c}
\noalign{\smallskip}
\hline
\noalign{\smallskip}
\multirow{2}{*}{Index} & \multirow{2}{*}{\textit{IoU\_threshold}} & \multirow{2}{*}{\textit{OR\_threshold}} & \multirow{2}{*}{\textit{num\_repeats}} & \multirow{2}{*}{\textit{stroke\_thickness}} & & \multicolumn{2}{c}{CBSC} & & \multicolumn{2}{c}{FrISS} & & \multirow{2}{*}{Avg} \\ \cline{7-8} \cline{10-11} & & & & & & AoN & S-IoU & & AoN & S-IoU & & \\
\noalign{\smallskip}
\hline
\noalign{\smallskip}

1 & 65\% & 60\% & 3 & 2 & & 74,17 & 88,19 & & 57,96 & 79,20 & & \textbf{74,88} \\
2 & 75\% & 45\% & 1 & 2 & & 73,53 & 87,56 & & 59,37 & 78,98 & & 74,86 \\
3 & 55\% & 60\% & 3 & 2 & & 74,17 & 88,19 & & 57,71 & 79,32 & & 74,85 \\
4 & 65\% & 75\% & 3 & 2 & & 73,56 & 87,94 & & 58,08 & 79,25 & & 74,71 \\
5 & 65\% & 70\% & 3 & 2 & & 73,56 & 87,94 & & 58,10 & 79,16 & & 74,69 \\
6 & 55\% & 55\% & 5 & 2 & & 74,07 & 88,11 & & 57,38 & 78,99 & & 74,64 \\
7 & 55\% & 70\% & 3 & 2 & & 73,56 & 87,94 & & 57,85 & 79,18 & & 74,63 \\
8 & 25\% & 60\% & 3 & 2 & & 74,40 & 88,47 & & 56,67 & 78,99 & & 74,63 \\
9 & 45\% & 60\% & 3 & 2 & & 74,17 & 88,27 & & 56,79 & 79,11 & & 74,59 \\
10 & 65\% & 70\% & 1 & 2 & & 73,43 & 88,02 & & 57,75 & 79,05 & & 74,56 \\
11 & 25\% & 75\% & 3 & 2 & & 73,79 & 88,20 & & 57,03 & 78,96 & & 74,49 \\
12 & 45\% & 75\% & 3 & 2 & & 73,56 & 88,00 & & 57,14 & 79,09 & & 74,45 \\
13 & 75\% & 70\% & 1 & 2 & & 72,69 & 87,64 & & 58,45 & 78,97 & & 74,44 \\
14 & 25\% & 70\% & 3 & 2 & & 73,79 & 88,20 & & 56,81 & 78,85 & & 74,41 \\
15 & 35\% & 75\% & 3 & 2 & & 73,34 & 87,97 & & 57,14 & 79,09 & & 74,38 \\
16 & 75\% & 50\% & 1 & 2 & & 72,89 & 87,61 & & 58,32 & 78,58 & & 74,35 \\
17 & 55\% & 50\% & 5 & 2 & & 73,76 & 87,84 & & 57,02 & 78,75 & & 74,34 \\
18 & 35\% & 75\% & 1 & 2 & & 73,21 & 88,05 & & 56,68 & 79,17 & & 74,28 \\
19 & 55\% & 50\% & 1 & 2 & & 73,63 & 87,89 & & 56,67 & 78,81 & & 74,25 \\
20 & 85\% & 75\% & 3 & 1 & & 73,23 & 87,41 & & 58,40 & 77,78 & & 74,21 \\ \noalign{\smallskip}
\hline
\noalign{\smallskip}
Lowest & 85\% & 50\% & 7 & 3 & & 66,97 & 83,00 & & 53,17 & 75,74 & & 69,72 \\
\noalign{\smallskip}
\hline
\noalign{\smallskip}
\end{tabular}}
\caption{The top-performing hyperparameter combinations for the post-processing module are presented in descending order.}
\label{table:ablation_cavt_postprocessing}
\end{table*}

\begin{itemize}
    \item \textbf{\textit{IoU\_threshold:}} The threshold value determines the Intersection over Union (IoU) of stroke sequences to boxes. For each box, if the IoU between the box and the longest intersecting stroke sequence exceeds \textit{IoU\_threshold}, the sequence is assigned to that box. For the ablation study, we adjusted the threshold within a range of $25\%$ to $85\%$, increasing by $10\%$ increments.
    \item \textbf{\textit{OR\_threshold:}} This is the threshold value that determines the assignment of remaining stroke sequences to boxes. If the overlap ratio of the longest unassigned stroke sequence with its nearest box exceeds \textit{OR\_threshold}, the sequence is assigned to that box. For the ablation study, we set the threshold ranges from $30\%$ to $80\%$ in $5\%$ increments.
    \item \textbf{\textit{num\_repeats:}} This refers to the total number of iterations the post-processing module undergoes to complete the stroke assignment process. The post-processing module continues until stroke group assignments reach a stable state. However, this approach can increase runtime, so we limited the number of iterations to evaluate the impact of different repetition counts. We tested the effect of the \textit{num\_repeats} parameter with values of 1, 3, 5, 7, and 9.
    \item \textbf{\textit{stroke\_thickness:}} We assessed the effect of stroke line thickness by evaluating the \textit{stroke\_thickness} parameter with values of 1, 2, 3, and 4, where higher values correspond to thicker stroke lines in the scene.
\end{itemize}

Table \ref{table:ablation_cavt_postprocessing} illustrates the impact of each parameter, revealing that the best-performing hyperparameter combination includes these values: \textit{IoU\_threshold} set to 65\%, \textit{OR\_threshold} to 60\%, \textit{num\_repeats} to 3, and \textit{stroke\_thickness} to 2. As demonstrated, using a value for \textit{stroke\_thickness} different than 2 degrades performance by distorting the features of the sketches. The \textit{num\_repeats} parameter does not significantly affect performance when increased, indicating that the stroke assignment operation completes effectively within a few iterations, minimizing the need for extended runtime. Setting \textit{OR\_threshold} to a low percentage can lead to incorrect stroke assignments, as some strokes that should be labeled as separate objects are merged with other stroke sequences. Therefore, setting \textit{OR\_threshold} higher than 50\% generally results in better performance. A range of 55\%-65\% for \textit{IoU\_threshold} yields the best results. Lower \textit{IoU\_threshold} values can lead to incorrect stroke-to-box assignments, while higher values may prevent the accurate stroke assignment.

\subsection{Post-Processing Time \& Memory Footprint}
\label{sec:postprocessing_timeandmemory}

Our post-processor takes on average 345 milliseconds per scene on CPU and has the memory upper bound of 5 times the scene in vector format.

\section{Additional Details on RGB Coloring Technique}
\label{sec:details_rgbcoloring}

\begin{figure}[th!]
  \centering
   \includegraphics[width=0.9\linewidth]{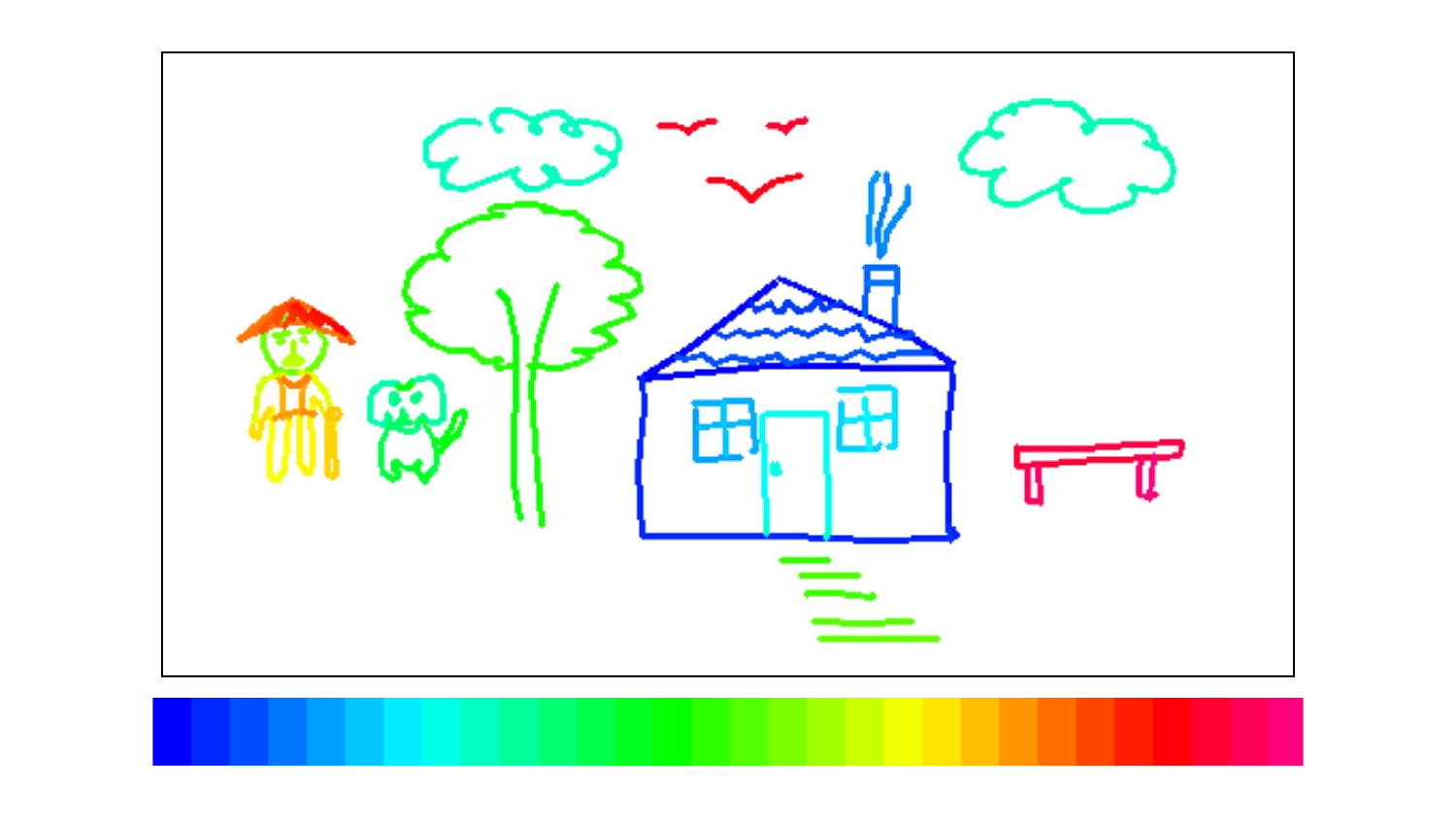}
  \caption{Sample scene sketch from the CBSC, which demonstrates the input for our object detector model. Each stroke within the scene is color-coded based on drawing order, utilizing a spectrum ranging from blue to red, as illustrated at the bottom.}
  \label{spectrum}
\end{figure}

\begin{table*}
\centering
\resizebox{0.9\textwidth}{!}{\begin{tabular}{lc|ccccc|ccccc|cccc}
\noalign{\smallskip}
\hline
\noalign{\smallskip}
\multirow{2}{*}{Model} & & & \multicolumn{3}{c}{Top-1 Accuracy} & & & \multicolumn{3}{c}{Top-3 Accuracy} & & & \multicolumn{3}{c}{Top-5 Accuracy} \\ \cline{4-6} \cline{9-11} \cline{14-16} \noalign{\smallskip} & & & CBSC & FrISS-QD & Avg. & & & CBSC & FrISS-QD & Avg. & & & CBSC & FrISS-QD & Avg. \\ 
\noalign{\smallskip}
\hline
\noalign{\smallskip}	
SketchR2CNN \cite{li2020sketch} & & & 63.04 & 48.65 & 55.85 & & & 71.57 & 59.13 & 65.35 & & & 74.12 & 63.06 & 68.59 \\
MGT \cite{xu2021multigraph} & & & 65.29 & 51.78 & 58.54 & & & 79.22 & 67.85 & 73.54	& & & 83.63 & 73.40 & 78.52	 \\
Sketchformer \cite{ribeiro2020sketchformer} & & & 65.88 & 52.82 & 59.35 & & & 80.81 & 66.57 & 73.69 & & & 85.69 & 71.36 & 78.53 \\
Inception-V3 \cite{szegedy2016rethinking} & & & \textbf{67.45} & \textbf{55.48} & \textbf{61.47} & & & \textbf{82.84} & \textbf{70.27} & \textbf{76.56} & & & \textbf{86.04} & \textbf{74.62} & \textbf{80.33} \\
\noalign{\smallskip}
\hline
\noalign{\smallskip}
\end{tabular}}
\caption{Analysis on state-of-the-art single sketch classifiers}
\label{table:classifiers_analysis}
\end{table*}

We adopted an RGB coloring technique to maintain a 3-channel input and values ranging from 0 to 255 for the detector. In our design, the neighboring strokes are represented with colors closer in the spectrum that spans from blue to red. Therefore, the strokes of the same object are expected to contain similar colors. Although a single object may not be entirely drawn in one stroke sequence, individual sequences are expected to exhibit consistent patterns. Besides the shape and distance of strokes, we expect our detector to recognize groups of consecutively sketched strokes. An illustrative example of a scene colored according to stroke order is given in Figure \ref{spectrum}.

\section{Additional Analysis on External Classifiers}
\label{sec:external_classifiers}

To develop a CNN-based sketch classifier, I first train several models, including Inception-V3 \cite{szegedy2016rethinking}, VGG19 \cite{simonyan2014very}, ResNet18 \cite{he2016deep}, ResNet50 \cite{he2016deep}, MobileNet-V3 \cite{howard2019searching}, and MobileNet-V2 \cite{sandler2018mobilenetv2}, using only the QuickDraw dataset. Afterward, I select the top three performing models and conduct further training by incorporating the FrISS training set along with QuickDraw. In both phases of the experiment, Inception-V3 consistently outperforms the other classifiers. Additionally, including the FrISS training set improves overall performance across both datasets. The results are summarized in Table \ref{table:cnn_classifier_analysis}. Based on these results, our pretrained Inception-V3 is selected as the external CNN-based classifier in our experiments.

\setlength{\tabcolsep}{4pt}
\begin{table}[ht!]
\begin{center}
\resizebox{\linewidth}{!}{\begin{tabular}{l|ccccccc}
\noalign{\smallskip}
\hline
\noalign{\smallskip}
\multirow{2}{*}{Model} & & \multicolumn{2}{c}{Train Dataset} & & \multicolumn{3}{c}{Accuracy} \\ \cline{3-4} \cline{6-8} \noalign{\smallskip} & & QD & FrISS & & CBSC & FrISS-QD & Avg. \\ 
\noalign{\smallskip}
\hline
\noalign{\smallskip}
Inception-V3 \cite{szegedy2016rethinking} & & \checkmark & & & 65.69 & 50.07 & \textcolor{green}{57.88} \\
VGG19 \cite{simonyan2014very} & & \checkmark & & & 64.02 & 50.69 & \textcolor{blue}{57.36} \\
ResNet18 \cite{he2016deep} & & \checkmark & & & 63.04 & 48.79 & \textcolor{red}{55.92} \\
ResNet50 \cite{he2016deep} & & \checkmark & & & 62.84 & 48.03 & 55.44 \\
MobileNetV3-Small \cite{howard2019searching} & & \checkmark & & & 61.37 & 46.85 & 54.11 \\
MobileNetV3-Large \cite{howard2019searching} & & \checkmark & & & 60.88 & 48.89 & 54.89 \\
MobileNet-V2 \cite{sandler2018mobilenetv2} & & \checkmark & & & 62.55 & 47.75 & 55.15 \\
\noalign{\smallskip}
\hline
\noalign{\smallskip}
Inception-V3 & & \checkmark & \checkmark & & 67.45 & 55.48 & \textbf{\textcolor{green}{61.47}} \\
VGG19 & & \checkmark & \checkmark 
 & & 65.98 & 55.24 & \textcolor{blue}{60.61} \\
ResNet18 & & \checkmark & \checkmark & & 67.65 & 53.11 & \textcolor{red}{60.38} \\
\noalign{\smallskip}
\hline
\noalign{\smallskip}
\end{tabular}}
\caption{The ablation study is performed to measure the effect of different backbone architectures and the effect of including the FrISS dataset in the training set. The highest average score is highlighted in \textcolor{green}{green}, the second highest in \textcolor{blue}{blue}, and the third highest in \textcolor{red}{red} for each aspect (i.e., backbone type and FrISS contribution).}
\label{table:cnn_classifier_analysis}
\end{center}
\end{table}
\setlength{\tabcolsep}{1.6pt}

I evaluate the performance of several state-of-the-art stroke-based sketch classifiers \cite{xu2021multigraph, li2020sketch, ribeiro2020sketchformer, szegedy2016rethinking}, and results are provided in Table \ref{table:classifiers_analysis}. The highest-performing transformer-based classifier, Sketchformer \cite{ribeiro2020sketchformer} is outperformed by our pretrained Inception-V3 \cite{szegedy2016rethinking}. To demonstrate the compatibility of CAVT with a stroke-based external classifier, Sketchformer is utilized in an end-to-end manner. 

\begin{figure*}[ht!]
  \centering
   \includegraphics[width=\linewidth]{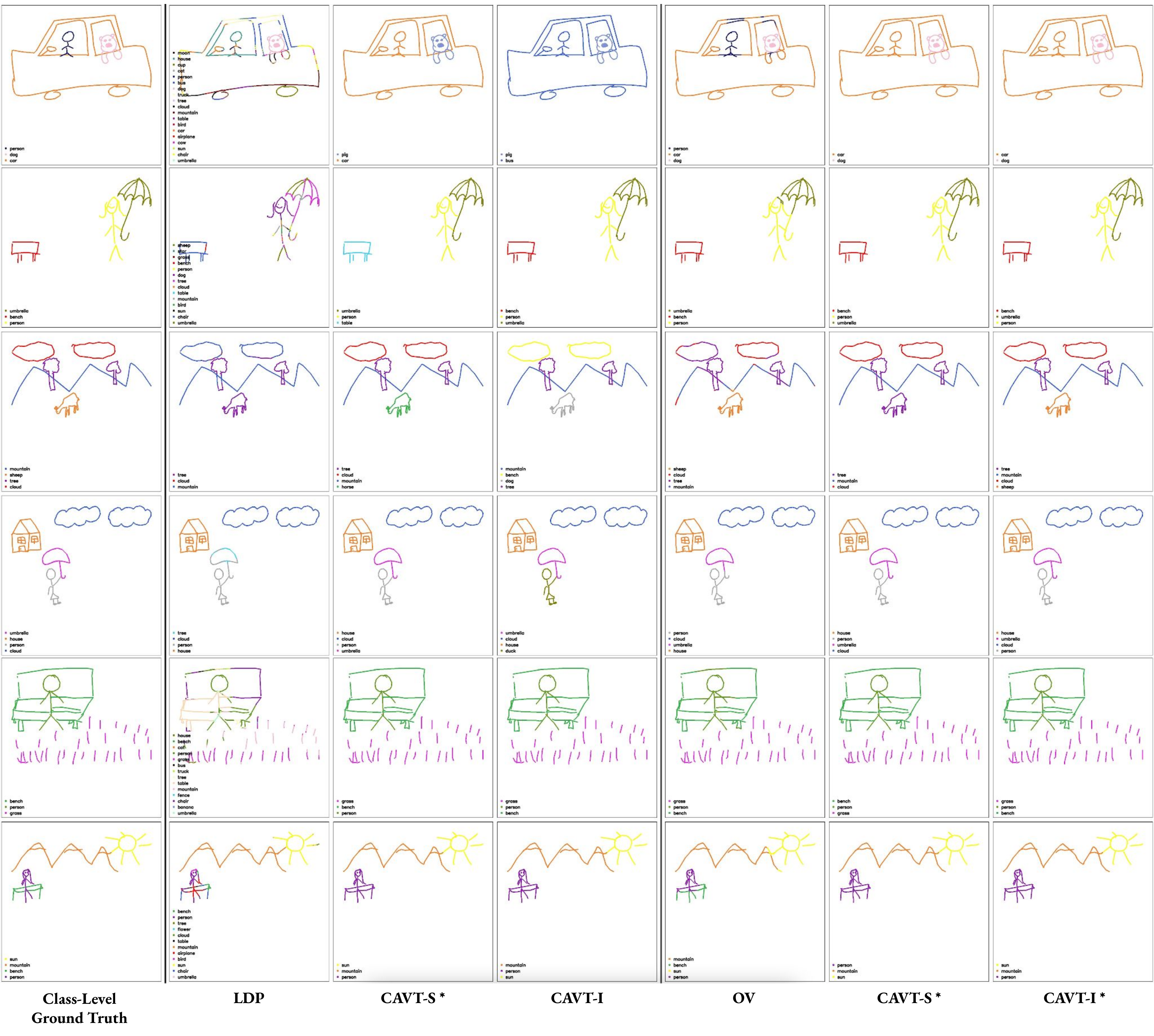}
  \caption{Visual comparison of our method with LDP \cite{ge2022exploring} and OV \cite{bourouis2023open} models, tested on FrISS dataset. We utilize CAVT with the external classifier Sketchformer \cite{ribeiro2020sketchformer} (CAVT-S) and our pre-trained Inception-V3 \cite{szegedy2016rethinking} (CAVT-I) in an end-to-end manner.}
  \label{visual-results-friss}
\end{figure*}

\begin{figure*}[ht!]
  \centering
   \includegraphics[width=\linewidth]{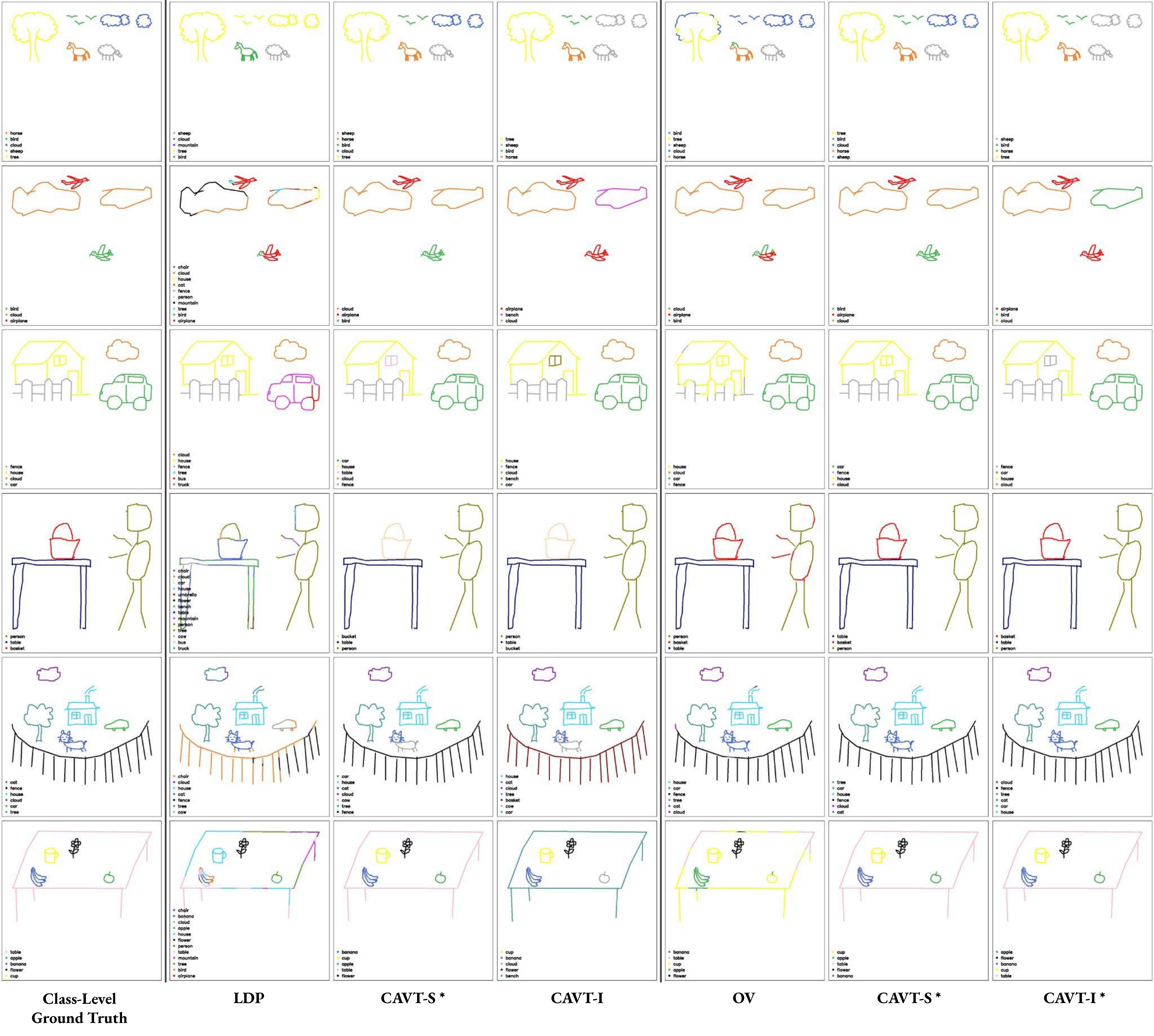}
  \caption{Visual comparison of our method with LDP \cite{ge2022exploring} and OV \cite{bourouis2023open} models, tested on CBSC dataset. We utilize CAVT with the external classifier Sketchformer \cite{ribeiro2020sketchformer} (CAVT-S) and our pre-trained Inception-V3 \cite{szegedy2016rethinking} (CAVT-I) in an end-to-end manner.}
  \label{visual-results-cbsc}
\end{figure*}

\begin{figure*}[ht!]
  \centering
   \includegraphics[width=0.6\textwidth]{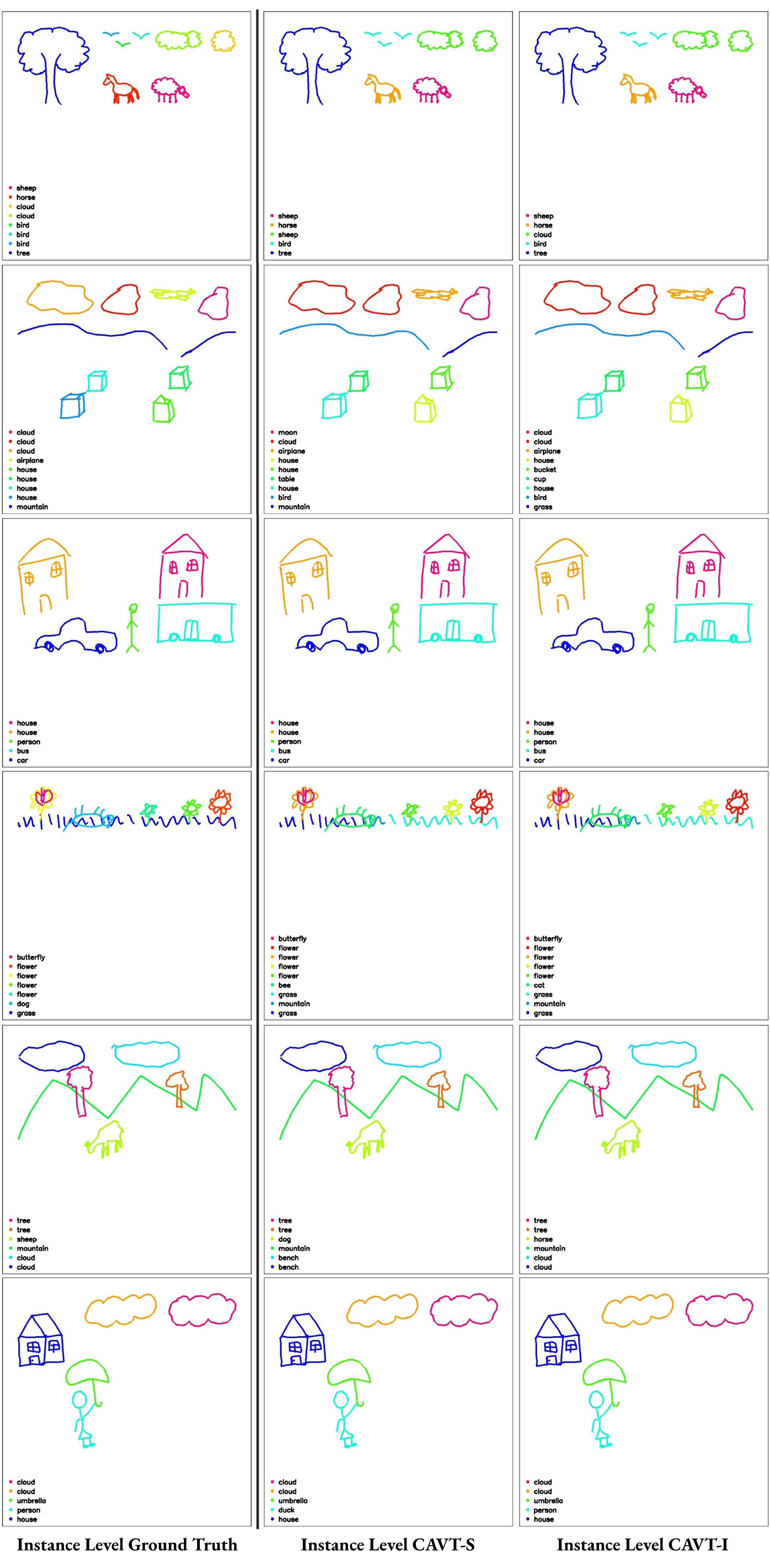}
  \caption{Instance-level visual results of CAVT in FrISS and CBSC datasets combined.}
  \label{inst-level-results}
\end{figure*}

\section{Additional Visual Results on Scene Sketch Semantic Segmentation}
\label{sec:additional_segmentation_results}

In Sec. \ref{sec:comparisons} of the main document, we provide a numerical comparison of the segmentation results obtained using our pipelines and two state-of-the-art methods: LDP \cite{ge2022exploring} and OV \cite{bourouis2023open}. Additionally, in Figure \ref{visual-results} from the main document, we present a visual comparison of our method against LDP and OV. Here, we provide additional visual results of our method against state-of-the-art models, assessed on FrISS and CBSC \cite{zhang2018context} datasets in Figures \ref{visual-results-friss} and \ref{visual-results-cbsc}, respectively. To visualize class-level segmentation results, we colored each pixel or stroke within the scene regarding its predicted object category. 

The additional visual outcomes depicted in Figures \ref{visual-results-friss} and \ref{visual-results-cbsc} demonstrate consistent segmentation results from both our primary pipelines (CAVT-S and CAVT-I) and its variant (CAVT-S* and CAVT-I). Therefore, we can observe that leveraging stroke representations of sketches and the temporal order of stroke sequences is a promising solution for the scene sketch segmentation problem. In some cases, although our class-agnostic approach successfully segments object instances, our adopted classifier may cause a performance drop due to its misclassification. For instance, in the 3rd row of Figure \ref{visual-results-friss}, our class-agnostic approach accurately segments the \textit{'sheep'} object. However, our adopted classifiers mislabel \textit{'sheep'} as \textit{'horse'} and \textit{'dog'}, thus impacting the segmentation results at the class level. This highlights the potential for our class-agnostic method's improved performance when paired with a classifier offering more accurate object class predictions. A similar issue is observed for the \textit{'cloud'} object in the 2nd row of Figure \ref{visual-results-cbsc}.

In addition to the class-level results, we share additional instance-level segmentation results in Figure \ref{inst-level-results}. In this figure, we can see that our pipelines successfully segment the objects from the same categories. While two houses are successfully differentiated in the 3rd row, the clouds are successfully detected and identified in the 6th row. However, there also exist some rare cases in which CAVT fails to segment (see individual birds and clouds in the 1st row).

\section{Additional Details on FrISS Dataset}
\label{sec:additional_details_friss}

\subsection{UI of Data Collection Web Application}
\label{sec:webui_friss}

\begin{figure}[t]
  \centering
   \includegraphics[width=\linewidth]{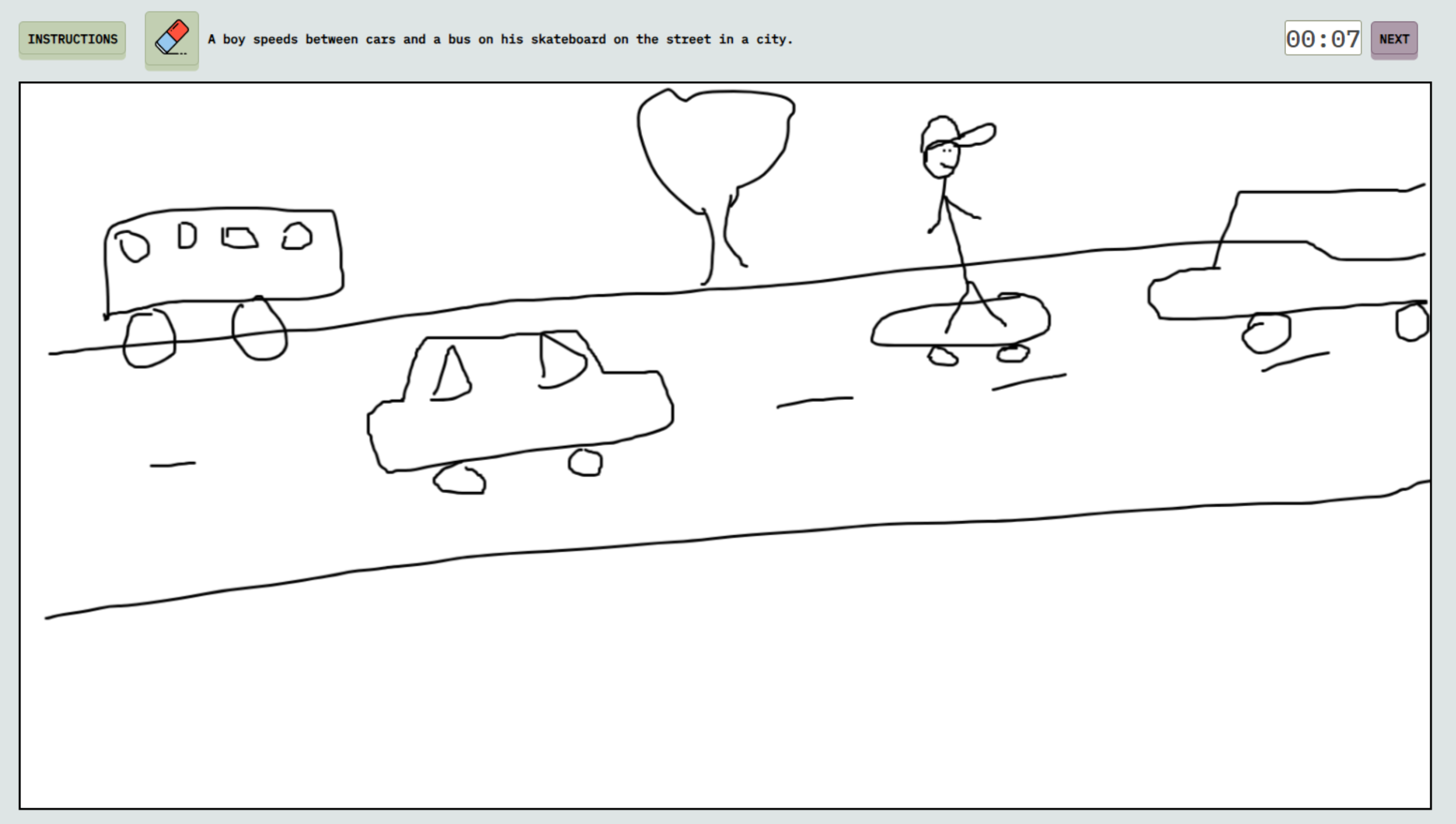}
\caption{The screenshot from the UI of the data collection web application during the drawing phase}
\label{ui_drawing}
\end{figure}

\begin{figure}[t]
\centering
\includegraphics[width=\linewidth]{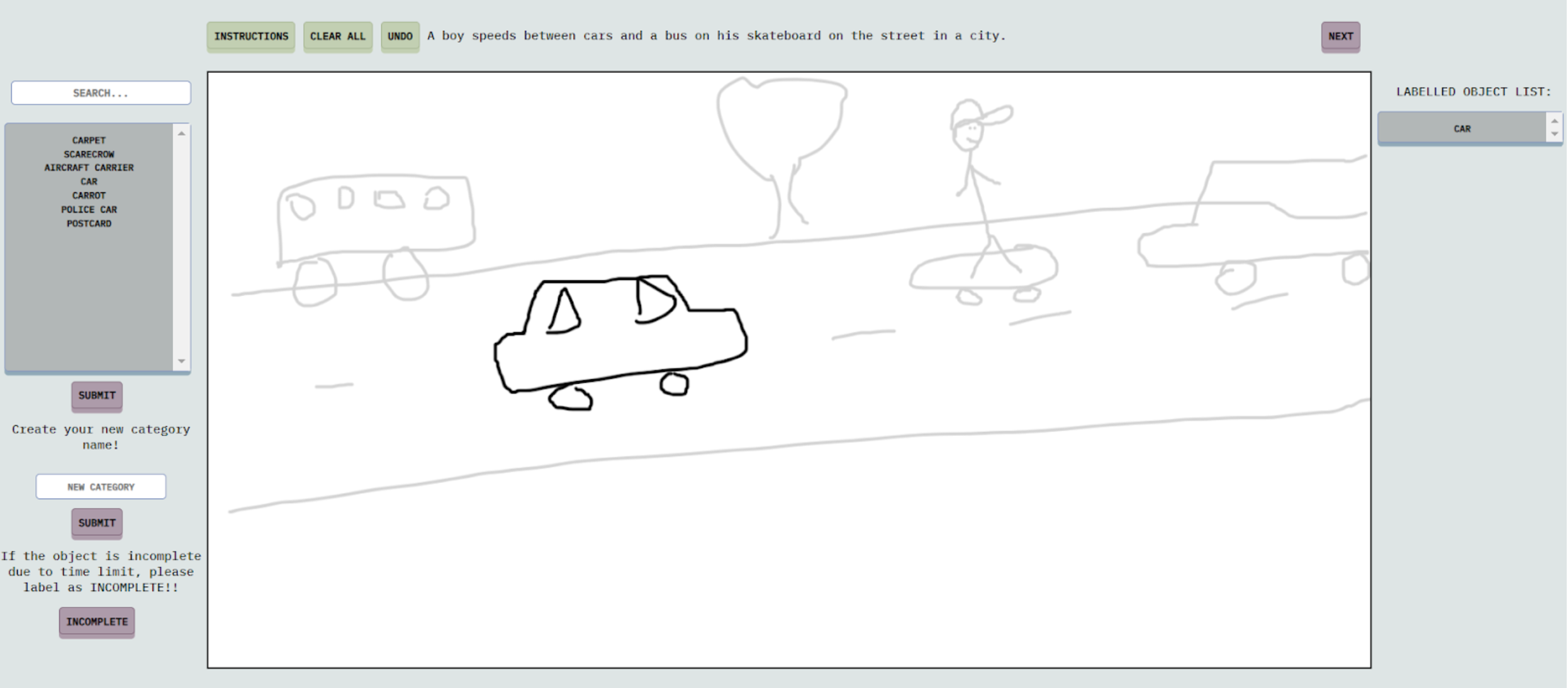}
\includegraphics[width=\linewidth]{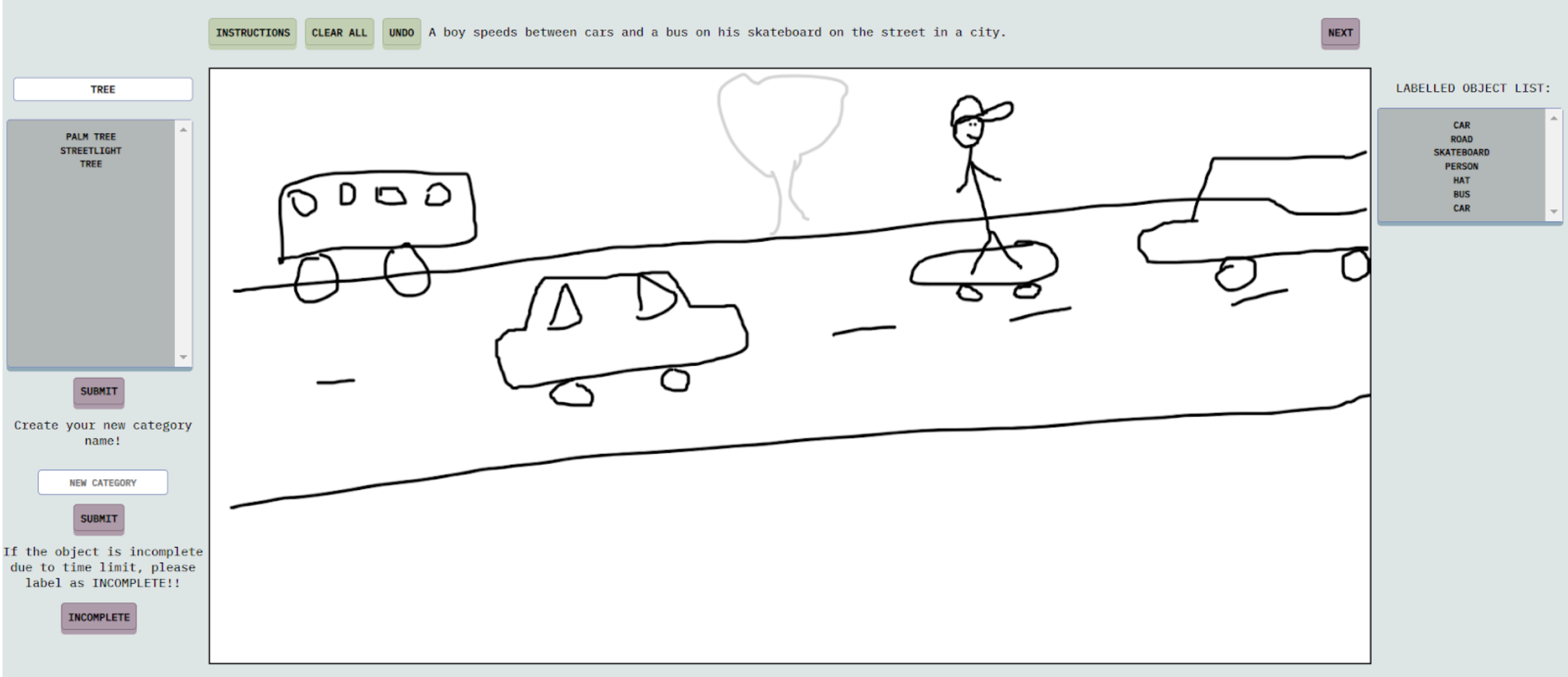}
\caption{The screenshot of data collection UI during the annotation phase. The upper image is taken while labeling the strokes corresponding to the initial object, \textit{'car'}. The lower image is taken before labeling the final drawn object, \textit{'tree'}. Annotated object classes are listed in the upper-right corner of the UI, in the order of labeling.}
\label{ui_annotation}
\end{figure}

In Sec. \ref{sec:dataset} of the main document, we provide a detailed discussion of our data collection process. In Figures \ref{ui_drawing} and \ref{ui_annotation}, we present visuals from the user interface of our data collection web application. As we discussed in the main document, our data collection consists of two distinct phases: sketch collection and sketch annotation. Figure \ref{ui_drawing} provides an example of the sketch collection phase, where participants are tasked with illustrating a scene within a time frame of 1.5 minutes, using a provided text description as a reference. Each participant sequentially draws 10 distinct scene sketches by referring to the corresponding descriptions. Upon completing the sketch collection phase, participants proceed to the second phase, where they annotate their previously drawn sketches.

During the annotation phase, depicted in Figure \ref{ui_annotation}, selected strokes turn from \textit{'gray'} to \textit{'black'} and participants assign a category to each stroke that turns into \textit{'black'}. The annotation process continues until each object instance within the scene is labeled (i.e., each stroke turns into \textit{'black'}). In the process of assigning categories, participants have the option to select from a predetermined list or introduce new categories by entering them into a designated text box (see Figure \ref{ui_annotation}). The predetermined list includes all QuickDraw \cite{ha2018a} classes and additional well-known categories not included in QuickDraw but likely to be sketched by participants (e.g., balloon, plate, carpet). This list is provided to ease the labeling process. Finally, strokes that are labeled as incompletely sketched or unrecognizable are marked as \textit{'incomplete'} and excluded from the dataset. Upon acceptance, we will release our data collection web application to the public.

\begin{figure*}[th!]
  \centering
   \includegraphics[width=\linewidth]{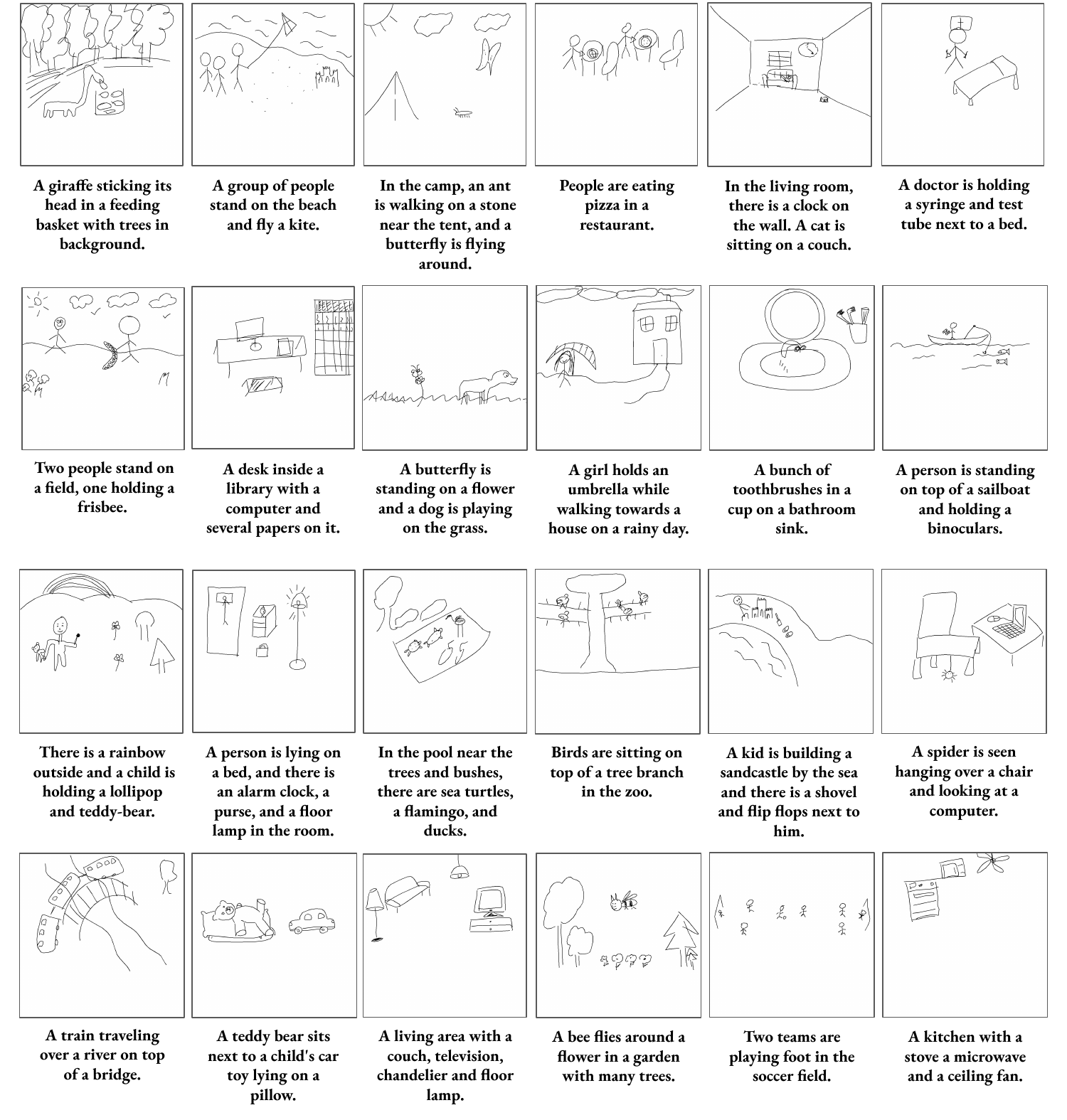}
  \caption{Sample scene sketches from our FrISS dataset paired with their textual scene descriptions}
  \label{friss-dataset-samples}
\end{figure*}

\begin{figure*}[t]
  \centering
   \includegraphics[width=\linewidth]{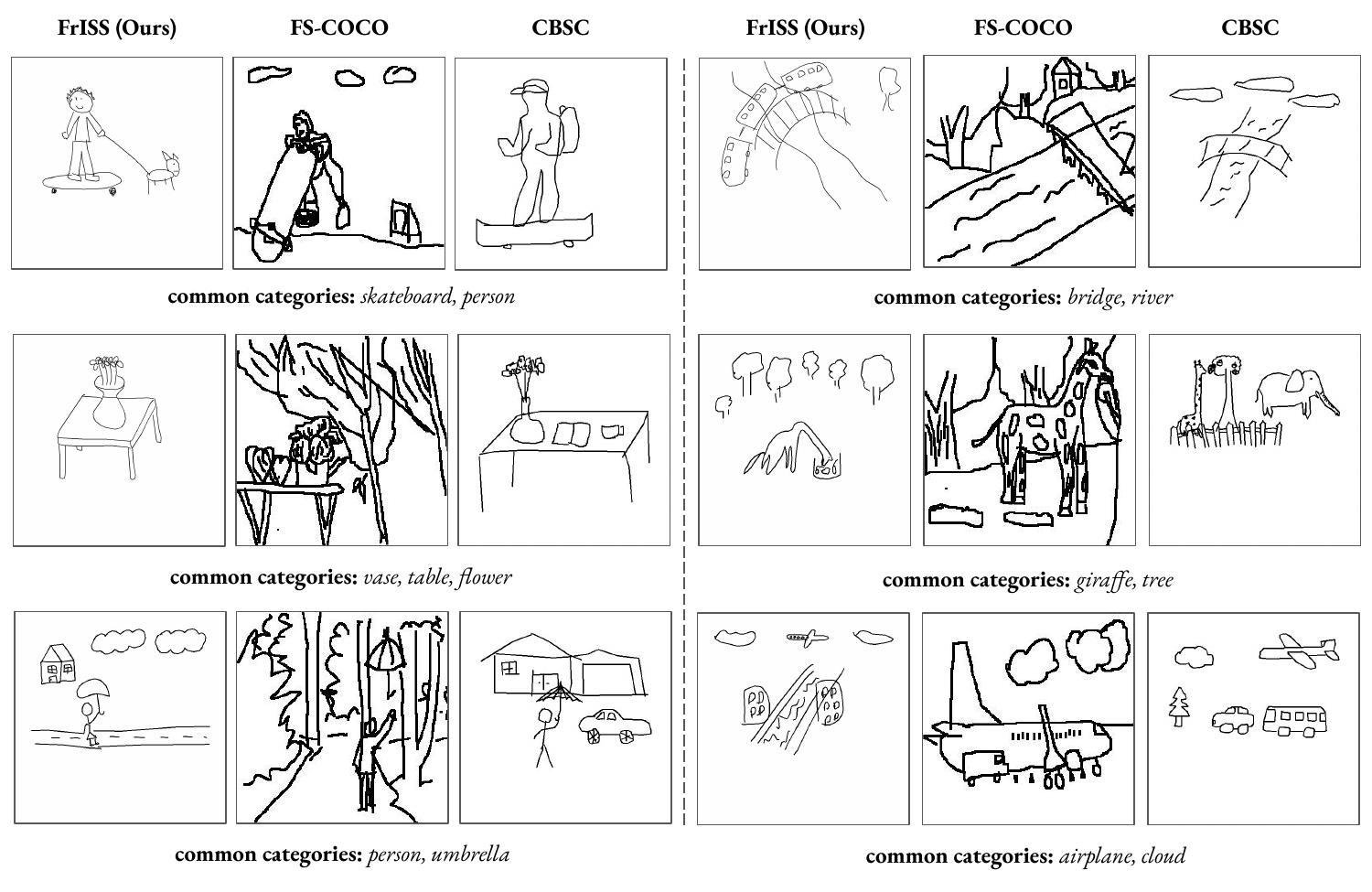}
  \caption{A comparison of scene sketches from FrISS with those from FS-COCO \cite{chowdhury2022fs} and CBSC \cite{zhang2018context}. The visuals are selected to ensure that each set of scene sketches shares at least two object categories in common, with the common classes listed below each group of three.}
  \label{dataset-comparison}
\end{figure*}

\begin{table*}
\centering
\resizebox{0.95\textwidth}{!}{\begin{tabular}{lllc}
\noalign{\smallskip}
\hline
\noalign{\smallskip}
Context & Scene Description & Expected Objects & COCO Img Id \\ 
\noalign{\smallskip}
\hline
\noalign{\smallskip}
bathroom & In the bathroom, there is a toilet, a bathtub, and a hair dryer. & toilet, bathtub, hair dryer & - \\
beach & A group of people stand on the beach and fly a kite. & person, kite, beach & 92478 \\
outdoor & A girl is standing next to a stop sign with an umbrella in her hand. & person, umbrella, stop sign	& - \\
garden & Four sheep are eating grass, and a child is approaching them. & person, sheep, grass & - \\
laboratory & A computer workstation with a printer, computer, mouse, and keyboards. & printer, computer, mouse, keyboard & 102609 \\
park & A skateboarder with a hat is riding his skateboard to walk his dog. & person, skateboard, dog, hat & 304173 \\
living room & A child eats ice cream and his eyeglasses fall on the carpet. & person, ice cream, carpet, eyeglasses	& - \\
hospital & A doctor is holding a syringe and test tube.	& person, syringe, test tube, bed & - \\
\noalign{\smallskip}
\hline
\noalign{\smallskip}
\end{tabular}}
\caption{Sample scene descriptions paired with the expected objects to be drawn by participants during the drawing phase of FrISS. The corresponding real-life image id is provided if the textual description is taken from the MS COCO dataset \cite{lin2014microsoft}.}
\label{table:scene_descriptions}
\end{table*}

\begin{figure*}[t]
  \centering
   \includegraphics[width=\linewidth]{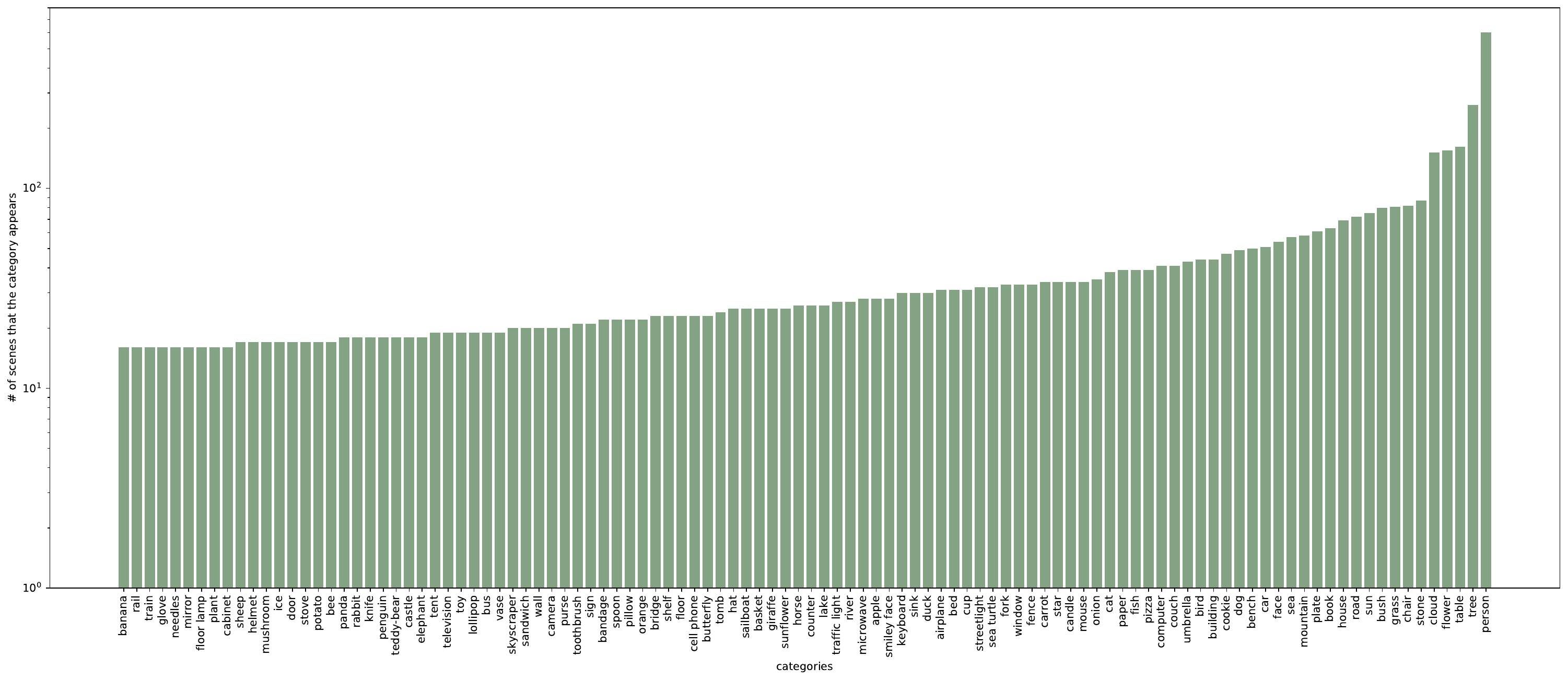}
\caption{The visualization of the number of scene sketches that each object category appears in. For visualization purposes, we selected the categories that have more than 15 appearances in the FrISS dataset.}
\label{categories_scene_counts}
\end{figure*}

\begin{figure}[t]
  \centering
   \includegraphics[width=\linewidth]{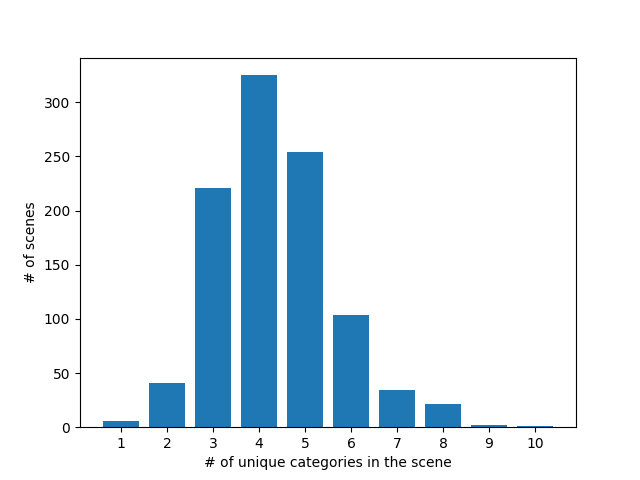}
\caption{The visualization of the number of unique object categories per scene in the FrISS dataset}
\label{categories_per_scene}
\end{figure}

\subsection{Visual Comparison of FrISS to Other Datasets}

In Sec. \ref{sec:datasetstatistics} of the main document, Table \ref{table:datasets_statistics} provides a statistical comparison of various scene sketch datasets, focusing on category, object, and stroke counts per sketch. Among these datasets provided in Table \ref{table:datasets_statistics}, CBSC \cite{zhang2018context}, FS-COCO \cite{chowdhury2022fs}, and SFSD \cite{zhang2023stroke} contain free-hand scene sketches stored in vector format. In Figure \ref{dataset-comparison}, we provide a detailed visual comparison between FrISS and these datasets. However, we could only share the visual comparisons between CBSC and FS-COCO, as SFSD is not publicly available. Additionally, we include extra sample scene sketches from FrISS along with their corresponding textual scene descriptions in Figure \ref{friss-dataset-samples}.

CBSC \cite{zhang2018context} and FS-COCO \cite{chowdhury2022fs} are collected under similar conditions: participants are permitted multiple drawing attempts, with an average completion time of 3 minutes per scene. In contrast, we imposed a drawing time limit of 1.5 minutes for each scene in our dataset, allowing redraw attempts only within this constrained timeframe, without permitting complete redraws. As depicted in Figure \ref{dataset-comparison}, our free-hand scene sketches exhibit significantly fewer strokes per object compared to those in FS-COCO. Furthermore, in the creation of FS-COCO, participants were presented with natural images as references during the drawing process. This results in scene sketches with similar object positions and postures as those in the referenced images. Conversely, the CBSC dataset was collected by instructing participants to quickly draw simple scene sketches that convey semantic meaning to humans, without any time restrictions. Our scene sketches demonstrate comparable object complexities to those in CBSC. However, while CBSC comprises 331 scene sketches covering 74 object categories, FrISS consists of 1K free-hand scene sketches, spanning a broader spectrum of object categories, totaling 403.

\subsection{Details of Textual Scene Descriptions}
\label{sec:descriptions_friss}

Scene descriptions are sourced either from the MS COCO dataset image captions \cite{lin2014microsoft} or manually created by us. Relying solely on MS COCO captions was insufficient to cover a wider range of object categories due to the dataset's limited variety. To ensure a broader representation, we aimed to include descriptions with at least three objects per scene, making sure the prompts were simple and drawable by individuals without professional drawing skills. 

To increase scene variety, most of the descriptions were manually constructed. We first gathered a list of environments likely to contain everyday objects. Then, we constructed scene descriptions featuring approximately 3 to 5 objects, ensuring they could be easily drawn within a specified time limit. In total, 180 unique scene descriptions were created, covering 403 object categories in FrISS. Table \ref{table:scene_descriptions} presents a subset of our scene descriptions along with their environments. The list of contexts is as follows: \textit{beach, zoo, sky, living room, ocean, kitchen, military base, stadium, concert hall, river, airport, hospital, jungle, graveyard, laboratory, camping site, restaurant, garden, gym, bedroom, gas station, battlefield, library, tower, school, cave, police station, space, museum, hotel, court, farm, hairdresser, park, bathroom, business center, music store, outdoor}.

\subsection{Detailed Analysis of FrISS}

Here, we provide additional analysis on our collected dataset in Figures \ref{categories_per_scene} and \ref{categories_scene_counts}. In Figure \ref{categories_per_scene}, we observe that the count of distinct object categories within a scene varies between 1 and 10, with a dominant accumulation between 3 and 6. Additionally, Figure \ref{categories_scene_counts} reveals that the most frequently occurring object categories in FrISS are person, tree, table, flower, and cloud, with the remaining categories distributed more balanced throughout the dataset. \\

\textbf{List of Categories in FrISS:} airplane, alarm clock, ambulance, ant, apple, arm, asparagus, axe, backpack, banana, bandage, barn, baseball, baseball bat, basket, basketball, bathtub, beach, bear, bed, bee, belt, bench, bicycle, binoculars, bird, birthday cake, blackberry, book, boomerang, bowtie, bracelet, bread, bridge, broom, bucket, bus, bush, butterfly, cake, calendar, camera, campfire, candle, cannon, canoe, car, carrot, castle, cat, ceiling fan, cell phone, cello, chair, chandelier, clarinet, clock, cloud, coffee cup, compass, computer, cookie, cooler, couch, cow, crab, crayon, crown, cruise ship, cup, dishwasher, dog, dolphin, donut, door, dresser, drill, drums, duck, dumbbell, elephant, eraser, eyeglasses, face, fan, fence, fire hydrant, fireplace, fish, flamingo, flashlight, flip flops, floor lamp, flower, fork, garden, giraffe, grapes, grass, guitar, hammer, hand, harp, hat, headphones, helmet, horse, hot air balloon, hot dog, hourglass, house, ice cream, key, keyboard, knife, ladder, laptop, leaf, light bulb, lighter, lighthouse, lightning, lion, lollipop, mailbox, map, microphone, microwave, moon, motorbike, mountain, mouse, mug, mushroom, necklace, ocean, octopus, onion, oven, palm tree, panda, pants, paper clip, pear, peas, pencil, penguin, picture frame, pig, pillow, pizza, police car, pond, pool, popsicle, potato, purse, rabbit, radio, rain, rainbow, rake, remote control, rhinoceros, river, sailboat, sandwich, saw, saxophone, school bus, scissors, screwdriver, sea turtle, see saw, shark, sheep, shoe, shovel, sink, skateboard, skull, skyscraper, sleeping bag, smiley face, snake, snorkel, snowflake, snowman, soccer ball, sock, spider, spoon, squirrel, stairs, star, steak, stereo, stop sign, stove, strawberry, streetlight, string bean, submarine, suitcase, sun, swan, swing set, syringe, t-shirt, table, teddy-bear, telephone, television, tennis racquet, tent, toilet, toothbrush, toothpaste, tractor, traffic light, train, tree, truck, trumpet, umbrella, vase, washing machine, watermelon, waterslide, wheel, windmill, wine bottle, wine glass, wristwatch, yoga*, zebra, anchor, bag, ball, balloon, barrier, baseball field, basketball hoop, bee nest, bell, billboard, board, bone, bottle, bowl, box, branch, building, button, cabinet, cable, cage, candy, carpet, cave, ceiling, cheese, chicken, cockroach, coconut, computer case, container, coral, counter, crosswalk, cupboard, curly hair, curtain, dagger, desk, dirt, dog collar, drain, drawer, earth, egg, exhibition, field, fish tank, fishing net, fishing rod, flag, floor, football field, footprint, fridge, frisbee, gas pump, gas station, glass, glass shard, glove, goal, gun, hair, hair dryer, hair tie, hammock, handcuffs, hanger, heart, hook, ice, jellyfish, kite, lake, lamp, light effect, marshmallow, meat, mirror, monitor, moon crater, mousepad, mud, museum, music note, necktie, needles, net, notebook, notes, orange, paddle, paper, path, pathway, peach, pepper, phone box, picnic rug, pipe, plant, plate, plug, present, printer, propeller, rail, restaurant, ribbon, road, rocket, roof, room, rope, ruler, safe, salt, sand, sandcastle, sausage, scarecrow, scarf, sea, sea fish, sea goggles, sea horse, sea shell, seagull, serum, shelf, shower head, sidewalk, sign, slide, smoke, soccer field, speaker, spider web, stage, stage lights, stand, staple, station, stick, stone, stool, strainer, street, suit, sunflower, sunglasses, surfboard, swim goggles, tape player, tennis court, test tube, toilet paper, tomb, tower, toy, traffic cone, trash bin, tray, tribune, turnstile, wall, walnut, water, weapon, wind, window, wing, wood. \\
Please note that in the FrISS dataset, \textit{yoga*} denotes the \textit{person} class. This mapping between the two classes is due to their visual similarity.

\subsection{Common Categories of FrISS and Other Datasets}

\begin{itemize}
    \item \textbf{List of common categories between FrISS and SKY-Scene \cite{ge2022exploring}:} airplane, apple, banana, bee, bench, bicycle, bird, butterfly, car, cat, chair, couch, cow, cup, dog, duck, flower, horse, house, mountain, pig, rabbit, sheep, strawberry, table, tree, truck, umbrella, wine bottle.
    \item \textbf{List of common categories between FrISS and SketchyScene \cite{zou2018sketchyscene}:} airplane, apple, banana, basket, bee, bench, bicycle, bird, bucket, bus, butterfly, car, cat, chair, cloud, couch, cow, cup, dog, duck, fence, flower, grass, horse, house, moon, mountain, pig, rabbit, sheep, star, streetlight, sun, table, tree, truck, umbrella, person.
    \item \textbf{List of common categories between FrISS and QuickDraw \cite{ha2018a}:} airplane, helicopter, alarm clock, clock, wristwatch, ambulance, firetruck, pickup truck, truck, leaf, van, apple, asparagus, onion, peas, potato, string bean, mushroom, backpack, banana, house, baseball, basketball, soccer ball, baseball bat, bear, panda, bed, bench, bicycle, bird, parrot, birthday cake, cake, blackberry, blueberry, grapes, pear, pineapple, strawberry, watermelon, book, bread, peanut, steak, bridge, broccoli, bus, school bus, bush, canoe, cruise ship, sailboat, speedboat, car, police car, carrot, cat, cell phone, chair, church, hospital, castle, cloud, coffee cup, cup, mug, computer, laptop, cooler, couch, cow, dog, donut, cookie, door, dresser, elephant, fence, fire hydrant, floor lamp, lantern, light bulb, flashlight, flower, fork, giraffe, hamburger, sandwich, horse, hot dog, house plant, jail, keyboard, knife, microwave, motorbike, mountain, mouse, ocean, oven, stove, dishwasher, washing machine, pillow, pizza, purse, rain, remote control, scissors, sheep, sink, skateboard, skyscraper, spoon, stairs, stop sign, suitcase, backpack, table, teddy-bear, television, tennis racquet, tent, toaster, toilet, toothbrush, traffic light, train, umbrella, vase, boomerang, basket, table, wine bottle, wine glass, person, zebra, stop sign, streetlight, hat, helmet, shoe, flip flops, eyeglasses, table, chandelier, ceiling fan, t-shirt, pants, dresser, pencil, eraser, grass, mountain, fence, river, sun, moon, star, snowflake, tree, palm tree
    \item \textbf{List of common categories between FrISS and CBSC \cite{zhang2018context}:} candle, bus, backpack, keyboard, car, camera, clock, mug, television, truck, banana, couch, elephant, flower, oven, pillow, cow, helmet, sheep, bridge, bench, table, spoon, horse, sandwich, bread, ladder, skateboard, tree, suitcase, bed, giraffe, house, fence, train, laptop, hat, bird, zebra, eyeglasses, fork, carrot, toilet, cat, person, airplane, baseball, bicycle, computer, basket, tent, stairs, chair, cell phone, river, cloud, knife, vase, umbrella, leaf, mountain, pizza, bucket, bear, cup, dog, bush, apple, key, cake, book, mouse, ocean.
    \item \textbf{List of common categories between FrISS and FS-COCO \cite{chowdhury2022fs}:} cloud, orange, cow, net, hot dog, car, couch, laptop, frisbee, road, chair, wine glass, roof, bed, horse, fork, knife, pizza, bird, river, sandwich, fire hydrant, floor, banana, apple, counter, backpack, bear, plate, mud, toothbrush, shoe, cup, airplane, umbrella, mountain, book, scissors, window, donut, bush, spoon, stairs, keyboard, vase, grass, wood, fence, bottle, kite, plant, mirror, traffic light, cat, door, oven, dog, truck, bus, zebra, toilet, bridge, skateboard, bench, table, dirt, bicycle, cage, giraffe, tent, tree, cake, picnic rug, bowl, stop sign, branch, house, sand, elephant, clock, cell phone, paper, skyscraper, baseball bat, carrot, suitcase, field, train, stone, sheep, surfboard, flower, hat, sea, person, tennis racquet.
\end{itemize}

\subsection{Ethical Considerations in Data Collection}

Our dataset contains free-hand scene sketches paired with their textual descriptions, audio clips of participants, and video recordings of drawing processes. During the drawing process, participants were asked to verbally explain their sketches in their native languages. At the beginning of the data collection, participants received detailed information regarding the following: the recording of their drawing screen in video format, the retention of their verbal descriptions as audio clips, and the potential release of their data in a research paper. Each participant was kindly requested to review and sign the consent form acknowledging our data collection procedures: \\

\textit{'I confirm that I have thoroughly read and understood the instructions. I hereby authorize the utilization of my anonymized data (i.e., drawings, video, and audio recordings) for scientific research purposes.'} \\

Participants who consented to our data collection terms were assigned a random ID and proceeded with the data collection process. Additionally, we provided a contact address to allow participants to confidentially address any concerns regarding the release of their data.

\end{document}